\documentclass{article}

\usepackage[toc,page,header]{appendix}
\usepackage{minitoc}
\noptcrule 

\usepackage{tocloft}
\PassOptionsToPackage{numbers, compress}{natbib}

\usepackage[final]{neurips_2021}

\usepackage[utf8]{inputenc} 
\usepackage[T1]{fontenc}    
\usepackage{hyperref}       
\usepackage{url}            
\usepackage{booktabs}       
\usepackage{amsfonts}       
\usepackage{nicefrac}       
\usepackage{microtype}      
\usepackage{xcolor}         
\usepackage{graphicx}
\usepackage{rotating}
\usepackage{amsmath}
\usepackage{multirow}
\usepackage{float}

\definecolor{forestgreen}{HTML}{228B22}

\title{An Image is Worth More Than a Thousand Words: Towards Disentanglement in the Wild}

\author{Aviv Gabbay  \qquad Niv Cohen \qquad Yedid Hoshen \\
School of Computer Science and Engineering \\
The Hebrew University of Jerusalem, Israel \\ ~\\
Project webpage: \textcolor{blue}{http://www.vision.huji.ac.il/zerodim}
}

\begin{document}

\doparttoc 
\faketableofcontents 

\maketitle

\begin{abstract}
  Unsupervised disentanglement has been shown to be theoretically impossible without inductive biases on the models and the data. As an alternative approach, recent methods rely on limited supervision to disentangle the factors of variation and allow their identifiability. While annotating the true generative factors is only required for a limited number of observations, we argue that it is infeasible to enumerate all the factors of variation that describe a real-world image distribution. To this end, we propose a method for disentangling a set of factors which are only partially labeled, as well as separating the complementary set of residual factors that are never explicitly specified. Our success in this challenging setting, demonstrated on synthetic benchmarks, gives rise to leveraging off-the-shelf image descriptors to partially annotate a subset of attributes in real image domains (e.g. of human faces) with minimal manual effort. Specifically, we use a recent language-image embedding model (CLIP) to annotate a set of attributes of interest in a zero-shot manner and demonstrate state-of-the-art disentangled image manipulation results. 
\end{abstract}

\section{Introduction}
\label{sec:introduction}
High-dimensional data (e.g. images) is commonly assumed to be generated from a low-dimensional latent variable representing the true factors of variation \cite{bengio2013representation, locatello2019challenging}. Learning to disentangle and identify these hidden factors given a set of observations is a cornerstone problem in machine learning, which has recently attracted much research interest \cite{higgins2016betavae,kim2018factorvae,chen2018tcvae,locatello2020fewlabels}. Recent progress in disentanglement has contributed to various downstream tasks as controllable image generation \cite{zhu2018visual}, image manipulation \cite{gabbay2020lord,gabbay2021overlord,wu2021stylespace} and domain adaptation \cite{peng2019domain}. Furthermore, disentangled representations pave the way for better interpretability \cite{hsu2017interpretable}, abstract reasoning \cite{van2019reasoning} and fairness \cite{creager2019flexiblyfair}.

A seminal study \cite{locatello2019challenging} proved that unsupervised disentanglement is fundamentally impossible without any form of inductive bias. While several different priors have been explored in recent works \cite{klindt2020towards,zimmermann2021contrastive}, the prominent approach is to introduce a limited amount of supervision at training time, i.e. assuming that a few samples are labeled with the true factors of variation \cite{locatello2020fewlabels}. There are two major limitations of such semi-supervised methods; (i) Manual annotation can be painstaking even if it is only required for part of the images (e.g. $100$ to $1000$ samples). (ii) For real-world data, there is no complete set of semantic and interpretable attributes that describes an image precisely. For example, one might ask: \textit{``Can an image of a human face be uniquely described with natural language?''}. The answer is clearly negative, as a set of attributes (e.g. age, gender, hair color) is far from uniquely defining a face.

Therefore, in this work we explore how 
to disentangle a few partially-labeled factors (named as \textit{attributes of interest}) in the presence of additional completely unlabeled attributes. We then show that we can obtain labels for these attributes of interest with minimal human effort by specifying their optional values as adjectives in natural language (e.g. "blond hair" or "wearing glasses"). Specifically, we use CLIP~ \cite{radford2021clip}, a recent language-image embedding model with which we annotate the training set images. As this model is already pretrained on a wide range of image domains, it provides rich labels for various visual concepts without any further manual effort in a zero-shot manner.

Nonetheless, leveraging general-purpose models as CLIP imposes a new challenge: among the attributes of interest, only part of the images are assigned to an accurate label. For this challenging disentanglement setting, we propose ZeroDIM, a novel method for Zero-shot Disentangled Image Manipulation. Our method disentangles a set of attributes which are only partially labeled, while also separating a complementary set of residual attributes that are never explicitly specified.

We show that current semi-supervised methods as Locatello et al. \cite{locatello2020fewlabels} perform poorly in the presence of residual attributes, while disentanglement methods that assume full supervision on the attributes of interest \cite{gabbay2020lord,gabbay2021overlord} struggle when only partial labels are provided. First, we simulate the considered setting in a controlled environment with synthetic data, and present better disentanglement of both the attributes of interest and the residual attributes. Then, we show that our method can be effectively trained with partial labels obtained by CLIP to manipulate real-world images in high-resolution.

\textbf{Our contributions are summarized as follows:} (i) Introducing a novel disentanglement method for the setting where a subset of the attributes are partially annotated, and the rest are completely unlabeled. (ii) Replacing manual human annotation with partial labels obtained by a pretrained language-image embedding model (CLIP). (iii) State-of-the-art results on synthetic disentanglement benchmarks and real-world image manipulation tasks.

\section{Related Work}
\label{sec:related_work}
\textbf{Semi-Supervised Disentanglement~} \citet{locatello2020fewlabels} investigate the impact of a limited amount of supervision on disentanglement methods and observe that a small number of labeled examples is sufficient to perform model selection on state-of-the-art unsupervised models. Furthermore, they show the additional benefit of incorporating supervision into the training process itself. In their experimental protocol, they assume to observe all ground-truth generative factors but only for a very limited number of observations. On the other hand, methods that do not require labels for all the generative factors \cite{cheung2014discovering,bouchacourt2018mlvae,denton2017drnet}, rely on full-supervision for the observed ones. In a seminal paper, \citet{kingma2014semi} study the setting considered in our work, where some factors are labeled only in a few samples, and the other factors are completely unobserved. However, the approach proposed in \cite{kingma2014semi} relies on exhaustive inference i.e. sampling all the possible factor assignments within the generative model. This exponential complexity inevitably limits the applicability of their approach for multi-attribute disentanglement. \citet{nie2020semistylegan} propose a semi-supervised StyleGAN for disentanglement learning by combining the StyleGAN architecture with the InfoGAN loss terms. Although specifically designed for real high-resolution images, it does not natively generalize to unseen images. For this purpose, the authors propose an extension named Semi-StyleGAN-fine, utilizing an encoder of a locality-preserving architecture, which is shown to be restrictive in a recent disentanglement study \cite{gabbay2021overlord}. Several other works \cite{klindt2020towards,zimmermann2021contrastive} suggest temporal priors for disentanglement, but can only be applied to sequential (video) data.

\textbf{Attribute Disentanglement for Image Manipulation~}
The goal in this task is to edit a distinct visual attribute of a given image while leaving the rest of the attributes intact. \citet{wu2021stylespace} show that the latent space spanned by the style channels of StyleGAN \cite{karras2019stylegan1,karras2020stylegan2} has an inherent degree of disentanglement which allows for high quality image manipulation. TediGAN \cite{xia2021towards} and StyleCLIP \cite{patashnik2021styleclip} explore leveraging CLIP \cite{radford2021clip} in order to develop a text-based interface for StyleGAN image manipulation. Such methods that rely on a pretrained unconditional StyleGAN generator are mostly successful in manipulating highly-localized visual concepts (e.g. hair color), while the control of global concepts (e.g. age) seems to be coupled with the face identity. Moreover, they often require manual trial-and-error to balance disentanglement quality and manipulation significance. Other methods such as LORD \cite{gabbay2020lord} and OverLORD \cite{gabbay2021overlord} allow to disentangle a set of labeled attributes from a complementary set of unlabeled attributes, which are not restricted to be localized (e.g. age editing). However, we show in the experimental section that the performance of LORD-based methods significantly degrades when the attributes of interest are only partially labeled.

\textbf{Joint Language-Image Representations~} Using natural language descriptions of images from the web as supervision is a promising direction for obtaining image datasets \cite{chen2015webly}. Removing the need for manual annotation opens the possibility of using very large datasets for better representation learning which can later be used for transfer learning \cite{joulin2016learning,quattoni2007learning,sariyildiz2020learning}. \citet{radford2021learning} propose learning a joint language-image representation with Contrastive Language-Image Pre-training (CLIP). The joint space in CLIP is learned such that the distance between the image and the text embedding is small for text-image pairs which are semantically related. This joint representation by CLIP was shown to have zero-shot classification capabilities using textual descriptions of the candidate categories. These capabilities were already used for many downstream tasks such as visual question answering \cite{gur2021cross}, image clustering \cite{cohen2021single}, image generation \cite{ramesh2021zero} and image manipulation \cite{xia2021towards,patashnik2021styleclip,bau2021paint}.

\section{Semi-Supervised Disentanglement with Residual Attributes}
\label{sec:method}
Assume a given set of images $x_1,x_2,...,x_n \in \mathcal{X}$ in which every image $x_i$ is specified precisely by a set of true generative factors. We divide these factors (or attributes) into two categories: 

\textbf{Attributes of Interest:} A set of semantic and interpretable attributes we aim to disentangle. We assume that we can obtain the labels of these attributes for a few training samples. We denote the assignment of these $k$ attributes of an image $x_i$ as $f^1_i, ..., f^k_i$.

\textbf{Residual attributes:} The attributes representing the remaining information needed to be specified to describe an image precisely. These attributes are represented in a single latent vector variable $r_i$.

As a motivational example, let us consider the task of disentangling human face attributes. The attributes of interest may include gender, age or hair color, while the residual attributes may represent the head pose and illumination conditions. Note that in real images, the residual attributes should also account for non-interpretable information e.g. details that relate to the facial identity of the person.

We assume that there exists an unknown function $G$ that maps $f^1_i, ..., f^k_i$ and $r_i$ to $x_i$:

\begin{equation}
\label{eq:formation}
    x_i = G(f^1_i, ..., f^k_i, r_i)
\end{equation}

A common assumption is that the true factors of variation are independent i.e. their density can be factorized as follows: $p(f^1, ..., f^k)=\prod^{k}_{j=1}p(f^j)$. Under this condition, the goal is to learn a representation that separates the factors of variation into independent components. Namely, a change in a single dimension of the representation should correspond to a change in a single generative factor. The representation of the residual attributes $r_i$, should be independent of all the attributes of interest. However, we only aim to learn a single unified representation for $r_i$, whose dimensions may remain entangled with respect to the residual attributes.

It should be noted that in real-world distributions such as real images, the true factors of variation are generally not independent. For example, the age of a person is correlated with hair color and the presence of facial hair. We stress that in such cases where the attributes are correlated, we restrict our attention to generating realistic manipulations: changing a target attribute with minimal perceptual changes to the rest of the attributes, while not learning statistically independent representations.

\subsection{Disentanglement Model}

\begin{figure}[t]
\begin{center}
\includegraphics[width=0.99\linewidth]{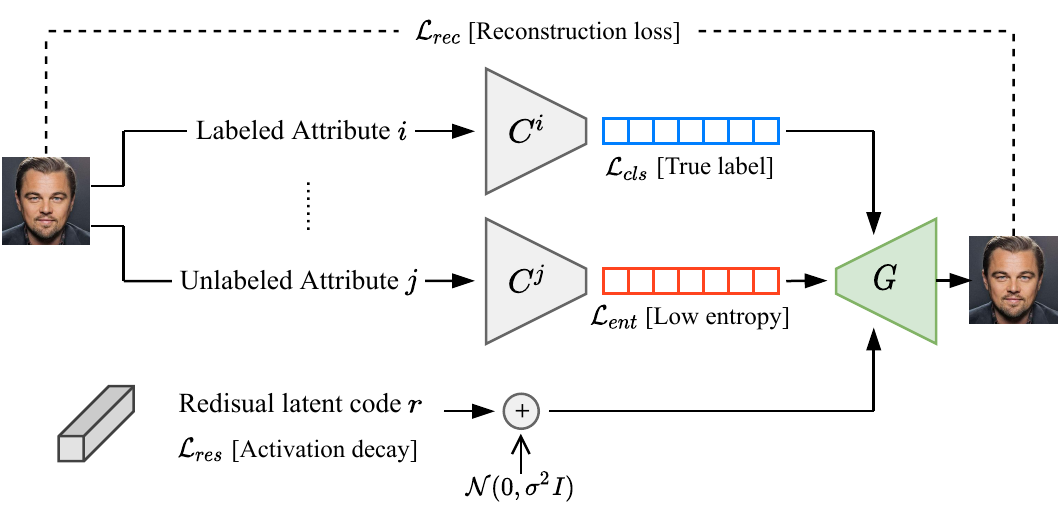}
\end{center}
\caption{A sketch of our method. Given a partially-labeled dataset, the attribute classifiers $\{C^i\}$ are trained with two complementary objectives: i) Predicting the true labels for the labeled samples. ii) Predicting a low-entropy estimation of the unknown labels. The entropy term constrains the information capacity and prevents the leakage of information not related to the specific attribute, encouraging disentanglement. The latent code $r$ (one per image) is regularized and optimized to recover the minimal residual information that is required for reconstructing the input image.}
\label{fig:method}
\end{figure}

Our model is aimed at disentangling the factors of variation for which at least \textit{some} supervision is given. The provided labels are indicated by the function $\ell$:

\begin{equation}
\label{eq:labeling_function}
\ell(i, j) = \begin{cases}
1 & \text{$f^j_i$ exists (attribute $j$ of image $i$ is labeled)}\\
0 & \text{otherwise} \\
\end{cases}
\end{equation}

For simplicity, we assume that each of the attributes is a categorical variable and train $k$ classifiers (one per attribute) of the form $C^j: \mathcal{X} \xrightarrow{} [m^j]$ where $m^j$ denotes the number of values of attribute $j$.

We optimize the classifiers using categorical cross-entropy loss, with the true labels when present:

\begin{equation}
\label{eq:loss_classification}
\mathcal{L}_{cls} = \sum^n_{i=1} \sum^k_{j=1} \ell(i, j) \cdot H \Big(\text{Softmax}\big(C^j(x_i)\big), f^j_i\Big)
\end{equation}

where $H(\cdot,\cdot)$ denotes cross entropy. For samples in which the true labels of the attributes of interest are not given, we would like the prediction to convey the relevant information, while not leaking information on other attributes. Therefore, we employ an entropy penalty $\mathcal{L}_{ent}$ that encourages the attribute value of each sample to be close to any of the one-hot vectors describing the known values, limiting its expressivity. The penalty is set over the classifier prediction, using the entropy $H(\cdot)$:

\begin{equation}
\label{eq:loss_entropy}
\mathcal{L}_{ent} = \sum^n_{i=1} \sum^k_{j=1} (1 - \ell(i, j)) \cdot  H \Big(\text{Softmax}\big(C^j(x_i)\big)\Big)
\end{equation}

To train the downstream part of our model, we set the value for each of the attributes of interest $j$ according to the label if given, or our classifier prediction otherwise: 

\begin{equation}
\label{eq:factor_inference}
\tilde{f}^j_i = \begin{cases}
f^j_i & \ell(i, j) = 1 \\
\text{Softmax\big($C^j(x_i)$\big)} & \ell(i, j) = 0 \\
\end{cases}
\end{equation}

To restrict the information of the attributes of interest from "leaking" into the residual code, we constrain the amount of information in the residual code as well. Naively, all the image attributes, labeled and residual alike, might be encoded in the residual representations. As we aim for the residual representations to contain only the information not available in the attributes of interest, we regularize the optimized latent codes with Gaussian noise and an activation decay penalty \cite{gabbay2020lord}:

\begin{equation}
\label{eq:loss_decay}
\mathcal{L}_{res} = \sum^n_{i=1} ||r_i||^2 ~~~~~~ r^{\prime}_i = r_i + \mu, \mu \sim \mathcal{N}(0, I)
\end{equation}

We finally employ a reconstruction loss to generate the target image:

\begin{equation}
\label{eq:loss_reconstruction}
\mathcal{L}_{rec} = \sum^n_{i=1} \phi(G(\tilde{f}^1_i, ..., \tilde{f}^k_i, r^{\prime}_i), x_i)
\end{equation}

where $\phi$ is a similarity measure between images, and is set to a mean-squared-error ($L_2$) loss for synthetic data and a perceptual loss for real images as suggested in \cite{gabbay2020lord}.

Our disentanglement model is trained from scratch in an end-to-end manner, optimizing a generator $G$, classifiers $C^1, ..., C^k$ and a residual latent code $r_i$ per image $x_i$, with the following objective: 
\begin{equation}
\label{eq:loss_disentanglement}
\min_{G,\{C^j\},\{r_i\}} ~~~ \mathcal{L}_{disen} = \mathcal{L}_{rec} + \lambda_{cls} \mathcal{L}_{cls} + \lambda_{ent} \mathcal{L}_{ent} + \lambda_{res} \mathcal{L}_{res} 
\end{equation}

A sketch of our architecture is visualized in Fig.~\ref{fig:method}.

\subsection{Implementation Details}
\paragraph{Latent Optimization} We optimize over the latents codes $r_i$ directly as they are \textit{not} parameterized by a feed-forward encoder. As discovered in \cite{gabbay2020lord}, latent optimization improves disentanglement over encoder-based methods. The intuition is that at initialization time: each $r_i$ is initialized i.i.d, by latent optimization and therefore is totally uncorrelated with the attributes of interest. However, a feed-forward encoder starts with near perfect correlation (the attributes could be predicted even from the output of a randomly initialized encoder). At the end of training using latent optimization, we possess representations for the residual attributes for every image in the training set. In order to generalize to unseen images, we then train a feed-forward encoder $E_r$ to infer the residual attributes by minimizing: $\mathcal{L}_{enc} = \sum^n_{i=1} \|E_r(x_i) - r_i\|^2$.

\paragraph{Warmup} The attribute classifiers predict the true labels when present and a low-entropy estimation of the label otherwise, to constrain the information capacity. As the classifiers are initialized randomly, we activate the entropy penalty ($\mathcal{L}_{ent}$) only after a fixed number of epochs during training.

More implementation details are provided in the Appendix.

\subsection{Experiments on Synthetic Datasets}
We first simulate our disentanglement setting in a controlled environment with synthetic data. In each of the datasets, we define a subset of the factors of variation as attributes of interest and the remaining factors as the residual attributes (the specific attribute splits are provided in the Appendix). For each attribute of interest, we randomly select a specific number of labeled examples (100 or 1000), while not making use of any labels of the residual attributes. As a complementary experiment, we also show state-of-the-art results in the semi-supervised setting of disentanglement with no residual attributes, which is the setting studied in \citet{locatello2020fewlabels} (see Appendix~\ref{app:semi_no_res}).

We experiment with four disentanglement datasets whose true factors are known: Shapes3D \cite{kim2018factorvae}, Cars3D \cite{reed2015deep}, dSprites \cite{higgins2016betavae} and SmallNORB \cite{lecun2004learning}. Note that partial supervision of 100 labels correspond to labeling 0.02\% of Shapes3D, 0.5\% of Cars3D, 0.01\% of dSprites and 0.4\% of SmallNORB.

\subsubsection{Baselines}

\paragraph{Semi-supervised Disentanglement} We compare against a semi-supervised variant of betaVAE suggested by \citet{locatello2020fewlabels} which incorporates supervision in the form of a few labels for each factor. When comparing our method, we utilize the exact same betaVAE-based architecture with the same latent dimension ($d=10$) to be inline with the disentanglement line of work. Our latent code is composed of two parts: a single dimension per attribute of interest (dimension $j$ is a projection of the output probability vector of classifier $C_j$), and the rest are devoted for the residual attributes $r_i$.

\paragraph{Disentanglement of Labeled and Residual Attributes} We also compare against LORD \cite{gabbay2020lord}, the state-of-the-art method for disentangling a set of labeled attributes from a set of unlabeled residual attributes. As LORD assumes full-supervision on the labeled attributes, we modify it to regularize the latent dimensions of the partially-labeled attributes in unlabeled samples to better compete in our challenging setting. See Appendix for a discussion on the relation of our method to LORD.

Note that we do not compare to unsupervised disentanglement methods \cite{higgins2016betavae,locatello2019challenging} as they can not compete with methods that incorporate supervision, according to the key findings presented in \cite{locatello2020fewlabels}.

\subsubsection{Evaluation}
We assess the learned representations of the attributes of interest using DCI \cite{eastwood2018dci} which measures three properties: (i) \textit{Disentanglement} - the degree to which each variable (or dimension) captures at most one generative factor. (ii) \textit{Completeness} - the degree to which each underlying factor is captured by a single variable (or dimension). (iii) \textit{Informativeness} - the total amount of information that a representation captures about the underlying
factors of variation. Tab.~\ref{tab:synthetic_quantitative} summarizes the quantitative evaluation of our method and the baselines on the synthetic benchmarks using DCI and two other disentanglement metrics: SAP \cite{kumar2017sapscore} and MIG \cite{chen2018tcvae}. It can be seen that our method learns significantly more disentangled representations for the attributes of interest compared to the baselines in both levels of supervision ($100$ and $1000$ labels). Note that while other disentanglement metrics exist in the literature, prior work has found them to be substantially correlated \cite{locatello2019challenging}.

Regarding the fully unlabeled residual attributes, we only require the learned representation to be informative of the residual attributes and disentangled from the attributes of interest. For evaluating these criteria, we train a set of linear classifiers, each of which attempts to predict a single attribute given the residual representations (using the available true labels). The representations learned by our method leak significantly less information regarding the attributes of interest. The entire evaluation protocol along with the quantitative results are provided in the Appendix for completeness. 

\subsection{Ablation Study}
\paragraph{Regularization Terms} We explore the contribution of the regularization terms introduced into our disentanglement objective (Eq.~\ref{eq:loss_disentanglement}). Training our model without the entropy penalty $\mathcal{L}_{ent}$ results in inferior disentanglement of the attributes of interest, while removing the residual codes regularization $\mathcal{L}_{res}$ leads to a leakage of information related to the attributes of interest into the residual representations. The quantitative evidence from this ablation study is presented in the Appendix.

\paragraph{Pseudo-labels} We consider a straightforward baseline in which we pretrain a classifier for each of the attributes of interest solely based on the provided few labels. 
We show in Tab.~\ref{tab:ablation_quantitative} that our method improves the attribute classification over these attribute-wise classifiers, implying the contribution of generative modeling to the discriminative-natured task of representation disentanglement \cite{atzmon2020causal}. Extended results from this study are provided in the Appendix.


\begin{table}[t]
  \caption{Evaluation on synthetic benchmarks using 1000 [or 100] labels per attribute of interest.}
  \label{tab:synthetic_quantitative}
  \centering
  \begin{tabular}{@{\hskip1pt}c@{\hskip6pt}lccccc@{\hskip1pt}}
    \toprule
    & & D & C & I & SAP & MIG \\
    \midrule
    
    \multirow{3}{*}{\rotatebox[origin=c]{90}{\scriptsize{\textbf{Shapes3D}}}} & \citet{locatello2020fewlabels} & 0.61 [0.03] & 0.61 [0.03] & 0.22 [0.22] & 0.05 [0.01] & 0.08 [0.02] \\
    & LORD \cite{gabbay2020lord} & 0.60 [0.54] & 0.60 [0.54] & 0.58 [0.54] & 0.18 [0.15] & 0.43 [0.42] \\
    & Ours & \textbf{1.00} [\textbf{1.00}] & \textbf{1.00} [\textbf{1.00}] & \textbf{1.00} [\textbf{1.00}] & \textbf{0.30} [\textbf{0.30}] & \textbf{1.00} [\textbf{0.96}] \\
    \midrule

    \multirow{3}{*}{\rotatebox[origin=c]{90}{\scriptsize{\textbf{Cars3D}}}} & \citet{locatello2020fewlabels} & 0.33 [0.11] & 0.41 [0.17] & 0.35 [0.22] & 0.14 [0.06] & 0.19 [0.04] \\
    & LORD \cite{gabbay2020lord} & 0.50 [0.26] & 0.51 [0.26] & 0.49 [0.36] & 0.19 [0.13] & 0.41 [0.20] \\
    & Ours & \textbf{0.80} [\textbf{0.40}] & \textbf{0.80} [\textbf{0.41}] & \textbf{0.78} [\textbf{0.56}] & \textbf{0.33} [\textbf{0.15}] & \textbf{0.61} [\textbf{0.35}] \\
    \midrule
    
    \multirow{3}{*}{\rotatebox[origin=c]{90}{\scriptsize{\textbf{dSprites}}}} & \citet{locatello2020fewlabels} & 0.01 [0.01] & 0.02 [0.01] & 0.13 [0.16] & 0.01 [0.01] & 0.01 [0.01] \\
    & LORD \cite{gabbay2020lord} & 0.40 [0.16] & 0.40 [0.17] & 0.44 [0.43] & 0.06 [0.03] & 0.10 [0.06] \\
    & Ours & \textbf{0.91} [\textbf{0.75}] & \textbf{0.91} [\textbf{0.75}] & \textbf{0.69} [\textbf{0.68}] & \textbf{0.14} [\textbf{0.13}] & \textbf{0.57} [\textbf{0.48}] \\
    \midrule
    
    \multirow{3}{*}{\rotatebox[origin=c]{90}{\scriptsize{\textbf{SmallNorb}}}} & \citet{locatello2020fewlabels} & 0.15 [0.02] & 0.15 [0.08] & 0.18 [0.18] & 0.02 [0.01] & 0.02 [0.01] \\
    & LORD \cite{gabbay2020lord} & 0.03 [0.01] & 0.04 [0.03] & 0.30 [0.30] & 0.04 [0.01] & 0.04 [0.02] \\
    & Ours & \textbf{0.63} [\textbf{0.27}] & \textbf{0.65} [\textbf{0.39}] & \textbf{0.53} [\textbf{0.45}] & \textbf{0.20} [\textbf{0.14}] & \textbf{0.40} [\textbf{0.27}] \\
    
    \bottomrule
  \end{tabular}
\end{table}

\begin{table}[b]
  \caption{Average attribute classification accuracy using 1000 [or 100] labels per attribute.}
  \label{tab:ablation_quantitative}
  \centering
  \begin{tabular}{lllll}
    \toprule
    & Shapes3D & Cars3D & dSprites &  SmallNORB  \\
    \midrule
    Pseudo-labels & 1.00	[0.84]	&	0.82	[0.46]	&   0.46	[0.28]	& 0.51	[0.38]	\\
    Ours & \textbf{1.00}	[\textbf{0.99}]	&	\textbf{0.85}	[\textbf{0.51}]	&	\textbf{0.68}	[\textbf{0.41}]  & \textbf{0.52}	[\textbf{0.39}] \\
    \bottomrule
  \end{tabular}
\end{table}

\section{ZeroDIM: Zero-shot Disentangled Image Manipulation}
\label{sec:clip_extension}

\begin{figure}[t]
\begin{center}
\includegraphics[width=1.0\linewidth]{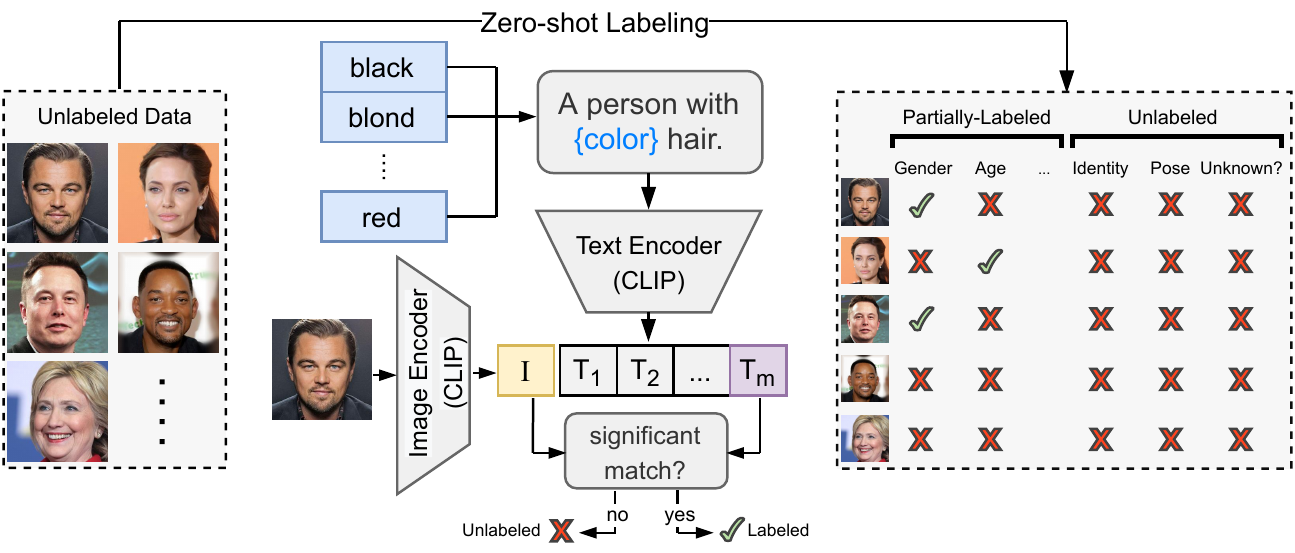}
\end{center}
\caption{A visualization of our zero-shot labeling for real-world unlabeled data. We define a list of attributes of interest and specify their optional values as adjectives in natural language e.g. \textit{blond} for describing \textit{hair color}. To annotate a specific attribute, we embed the images along with several natural sentences containing the candidate labels (prompt engineering \cite{radford2021clip}) into the joint embedding space of a pretrained CLIP. The images which correlate the most with a query are assigned to a label. If none of the labels are assigned, the image remains unlabeled.}
\label{fig:clip}
\vspace{-0.5em}
\end{figure}

In this section we make a further step towards disentanglement without manual annotation. We rely on the robustness of our method to partial-labeling which naturally fits the use of a general-purpose classification models such as CLIP \cite{radford2021clip}. CLIP is designed for zero-shot matching of visual concepts with textual queries. Our setting fits it in two aspects: (i) Only part of the attributes can be described with natural language. (ii) Only part of the images can be assigned to an accurate label. 

\subsection{Zero-shot Labeling with CLIP}
Our use of CLIP \cite{radford2021clip} to provide annotations is driven by a short textual input. We provide short descriptions, suggesting a few possible values of each attribute of interest e.g. \textit{hair color} can take one of the following: \textit{"red hair", "blond hair"} etc. Yet, there are two major difficulties which prevent us from labeling all images in the dataset: (i) Even for a specified attribute, not necessarily all the different values (e.g. different hair colors) can be described explicitly. (ii) The classification capabilities of a pretrained CLIP model are limited. This can be expected, as many attributes might be ambiguous (\textit{"a surprised expression" vs. "an excited expression"}) or semantically overlap (\textit{"a person with glasses" or "a person with shades"}). To overcome the "noisiness" in our labels, we set a confidence criterion for annotation: for each value we annotate only the best $K$ matching images, as measured by the cosine distance of embedding pairs. Fig.~\ref{fig:clip} briefly summarizes this labeling process.

\subsection{Experiments on Real Images}
\label{sec:experiment}

In order to experiment with real images in high resolution, we make two modifications to the method proposed in Sec.~\ref{sec:method}: (i) The generator architecture is adopted from StyleGAN2 \cite{karras2020stylegan2}, replacing the betaVAE decoder. (ii) An additional adversarial discriminator is trained with the rest of the modules for increased perceptual fidelity of the generated images, similarly to \cite{gabbay2021overlord}.

We demonstrate our zero-shot disentangled image manipulation on three different image domains: human faces (FFHQ \cite{karras2019stylegan1}), animal faces (AFHQ \cite{choi2020stargan}) and cars \cite{carsdataset}. The entire list of attributes of interest used in each dataset can be found in the Appendix.

\begin{figure*}[t]
\begin{center}
\begin{tabular}{@{\hskip0pt}c@{\hskip2pt}c@{\hskip2pt}c@{\hskip0pt}c@{\hskip0pt}c@{\hskip0pt}c@{\hskip0pt}c@{\hskip0pt}c@{\hskip0pt}c}

& Input & Kid & Asian & Gender & Glasses & Shades & Blond hair & Red hair \\

\begin{turn}{90} \footnotesize ~~ TediGAN \end{turn} &
\includegraphics[width=0.12\linewidth]{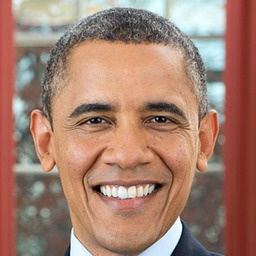} &
\includegraphics[width=0.12\linewidth]{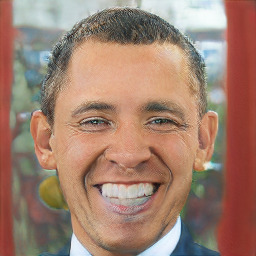} &
\includegraphics[width=0.12\linewidth]{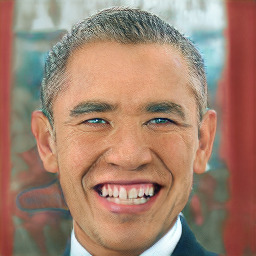} &
\includegraphics[width=0.12\linewidth]{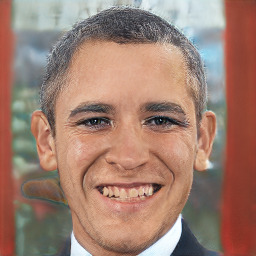} &
\includegraphics[width=0.12\linewidth]{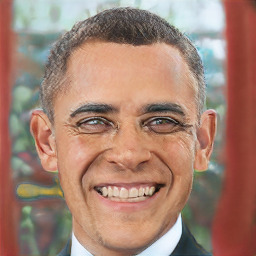} &
\includegraphics[width=0.12\linewidth]{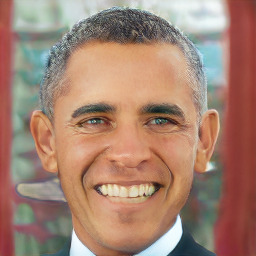} &
\includegraphics[width=0.12\linewidth]{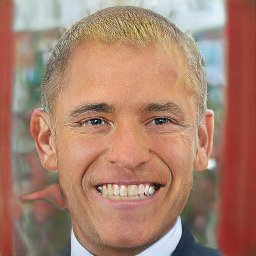} &
\includegraphics[width=0.12\linewidth]{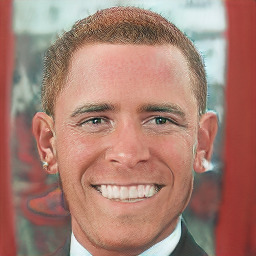} \\

\begin{turn}{90} \footnotesize ~ StyleCLIP- \end{turn} &
\includegraphics[width=0.12\linewidth]{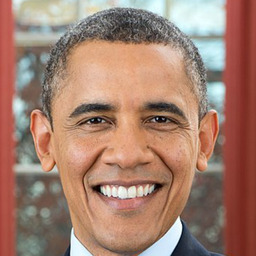} &
\includegraphics[width=0.12\linewidth]{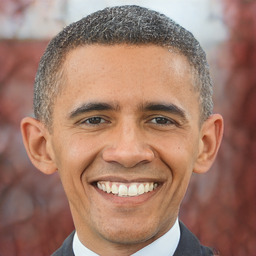} &
\includegraphics[width=0.12\linewidth]{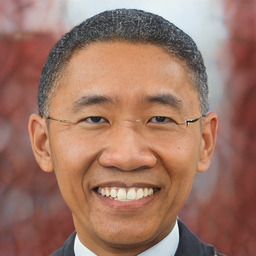} &
\includegraphics[width=0.12\linewidth]{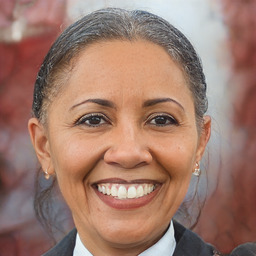} &
\includegraphics[width=0.12\linewidth]{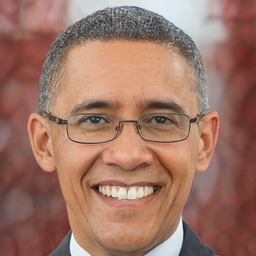} &
\includegraphics[width=0.12\linewidth]{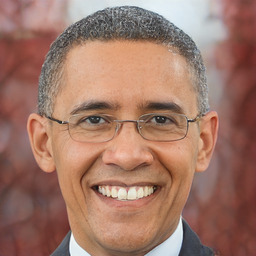} &
\includegraphics[width=0.12\linewidth]{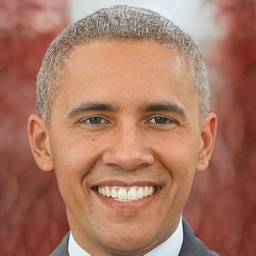} &
\includegraphics[width=0.12\linewidth]{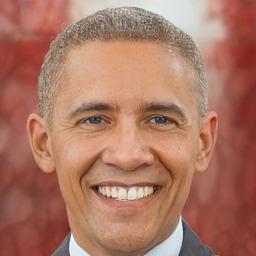} \\

\begin{turn}{90} \footnotesize ~ StyleCLIP+ \end{turn} &
\includegraphics[width=0.12\linewidth]{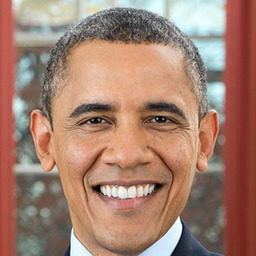} &
\includegraphics[width=0.12\linewidth]{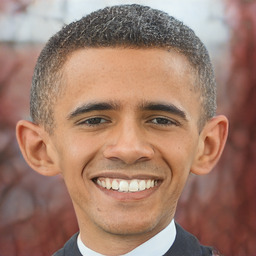} &
\includegraphics[width=0.12\linewidth]{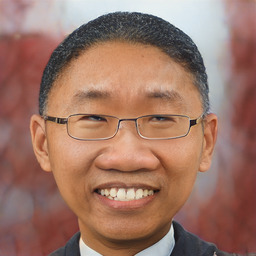} &
\includegraphics[width=0.12\linewidth]{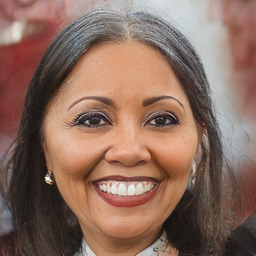} &
\includegraphics[width=0.12\linewidth]{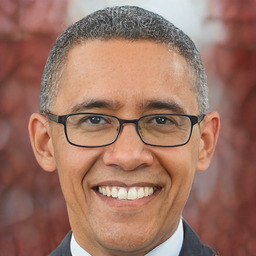} &
\includegraphics[width=0.12\linewidth]{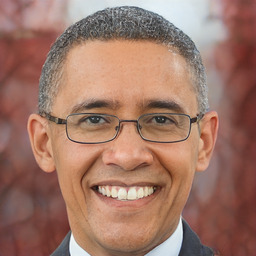} &
\includegraphics[width=0.12\linewidth]{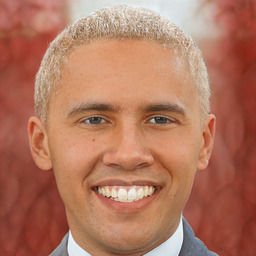} &
\includegraphics[width=0.12\linewidth]{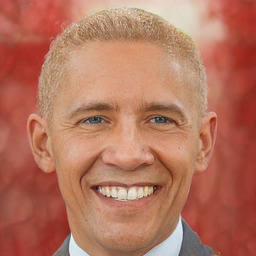} \\

\begin{turn}{90} ~~~ LORD \end{turn} &
\includegraphics[width=0.12\linewidth]{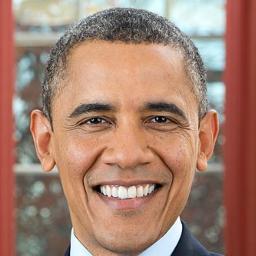} &
\includegraphics[width=0.12\linewidth]{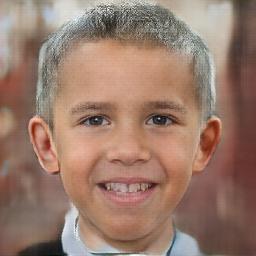} &
\includegraphics[width=0.12\linewidth]{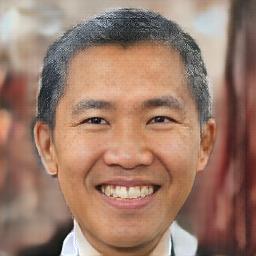} &
\includegraphics[width=0.12\linewidth]{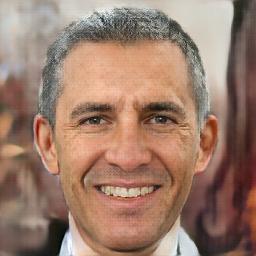} &
\includegraphics[width=0.12\linewidth]{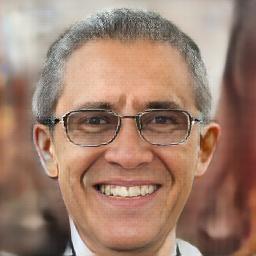} &
\includegraphics[width=0.12\linewidth]{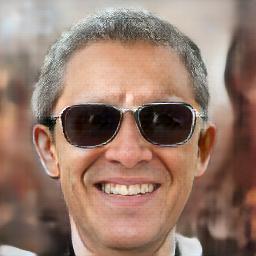} &
\includegraphics[width=0.12\linewidth]{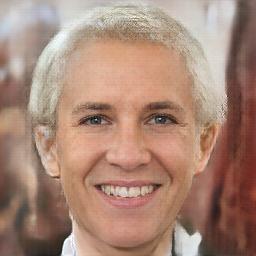} &
\includegraphics[width=0.12\linewidth]{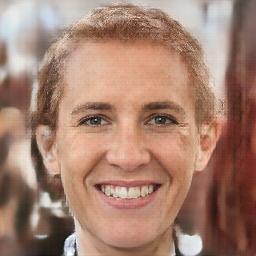} \\

\begin{turn}{90} ~~~~ \textbf{Ours} \end{turn} &
\includegraphics[width=0.12\linewidth]{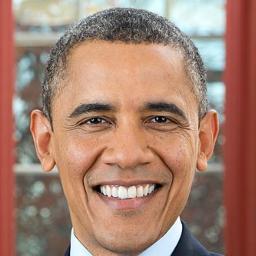} &
\includegraphics[width=0.12\linewidth]{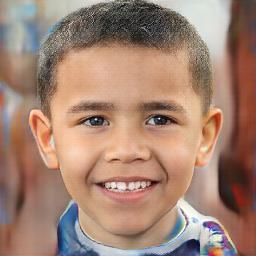} &
\includegraphics[width=0.12\linewidth]{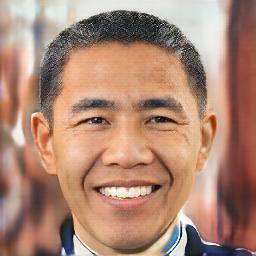} &
\includegraphics[width=0.12\linewidth]{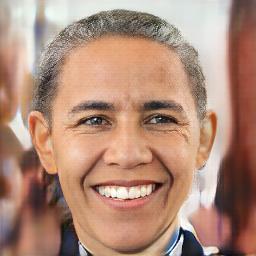} &
\includegraphics[width=0.12\linewidth]{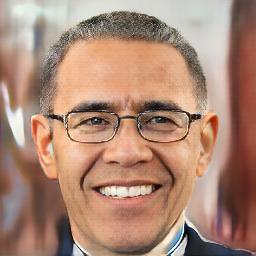} &
\includegraphics[width=0.12\linewidth]{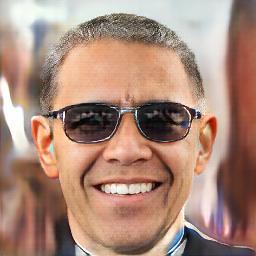} &
\includegraphics[width=0.12\linewidth]{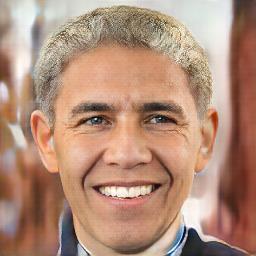} &
\includegraphics[width=0.12\linewidth]{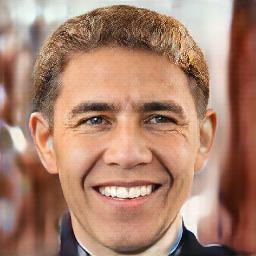} \\

\begin{turn}{90} \footnotesize ~~ TediGAN \end{turn} &
\includegraphics[width=0.12\linewidth]{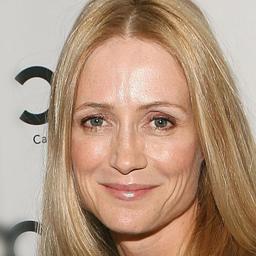} &
\includegraphics[width=0.12\linewidth]{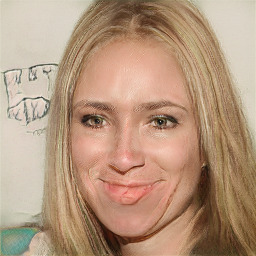} &
\includegraphics[width=0.12\linewidth]{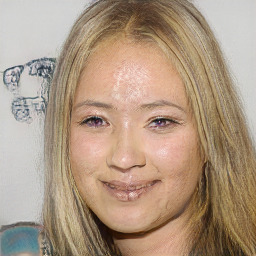} &
\includegraphics[width=0.12\linewidth]{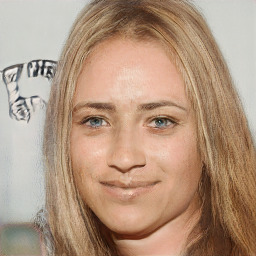} &
\includegraphics[width=0.12\linewidth]{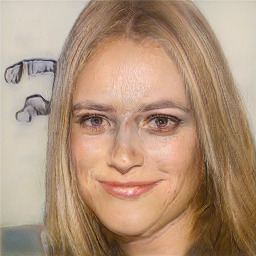} &
\includegraphics[width=0.12\linewidth]{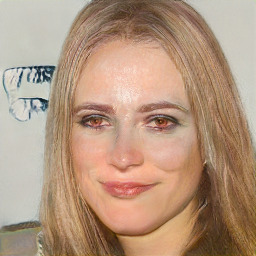} &
\includegraphics[width=0.12\linewidth]{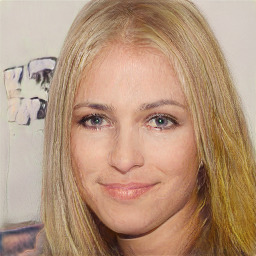} &
\includegraphics[width=0.12\linewidth]{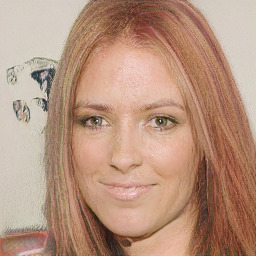} \\

\begin{turn}{90} \footnotesize ~ StyleCLIP-  \end{turn} &
\includegraphics[width=0.12\linewidth]{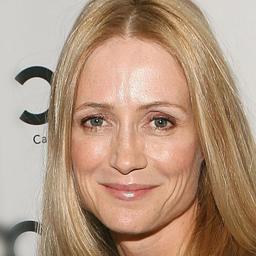} &
\includegraphics[width=0.12\linewidth]{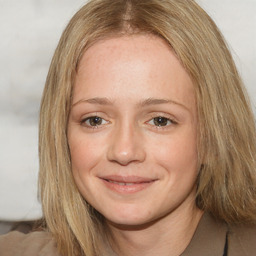} &
\includegraphics[width=0.12\linewidth]{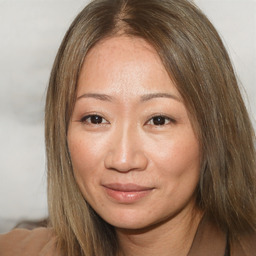} &
\includegraphics[width=0.12\linewidth]{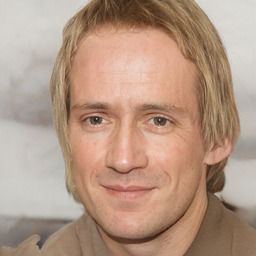} &
\includegraphics[width=0.12\linewidth]{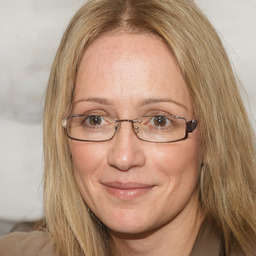} &
\includegraphics[width=0.12\linewidth]{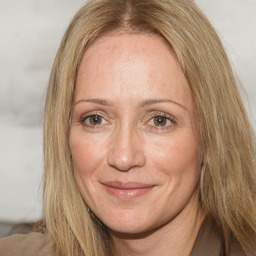} &
\includegraphics[width=0.12\linewidth]{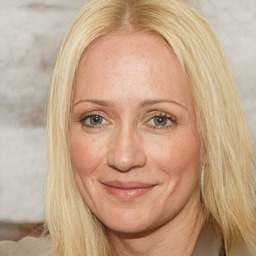} &
\includegraphics[width=0.12\linewidth]{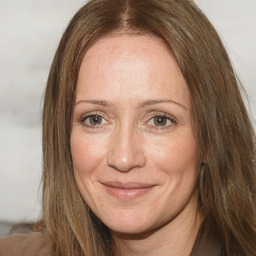} \\

\begin{turn}{90} \footnotesize ~ StyleCLIP+  \end{turn} &
\includegraphics[width=0.12\linewidth]{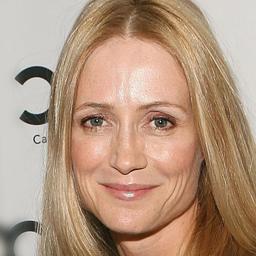} &
\includegraphics[width=0.12\linewidth]{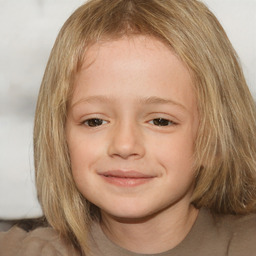} &
\includegraphics[width=0.12\linewidth]{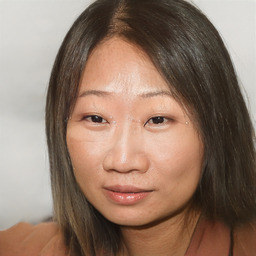} &
\includegraphics[width=0.12\linewidth]{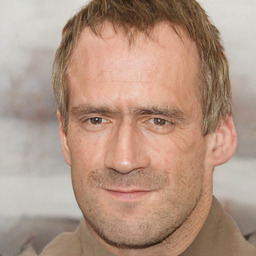} &
\includegraphics[width=0.12\linewidth]{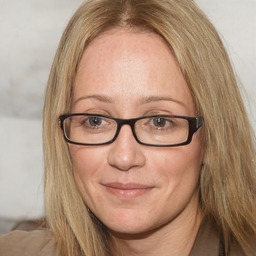} &
\includegraphics[width=0.12\linewidth]{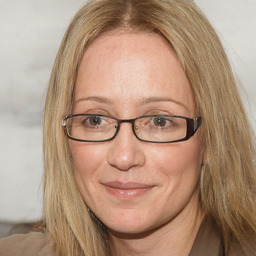} &
\includegraphics[width=0.12\linewidth]{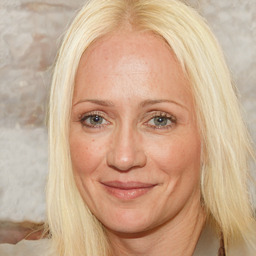} &
\includegraphics[width=0.12\linewidth]{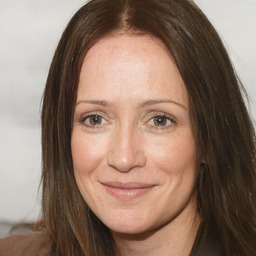} \\


\begin{turn}{90} ~~~~ \textbf{Ours} \end{turn} &
\includegraphics[width=0.12\linewidth]{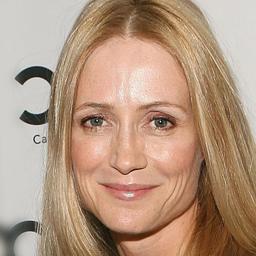} &
\includegraphics[width=0.12\linewidth]{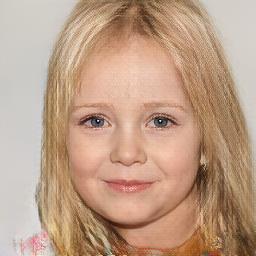} &
\includegraphics[width=0.12\linewidth]{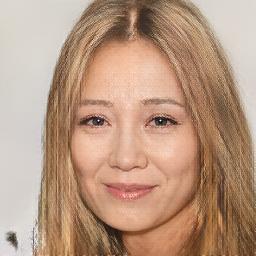} &
\includegraphics[width=0.12\linewidth]{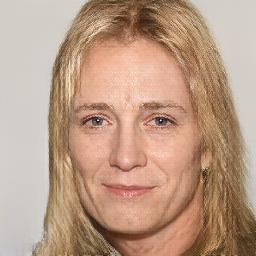} &
\includegraphics[width=0.12\linewidth]{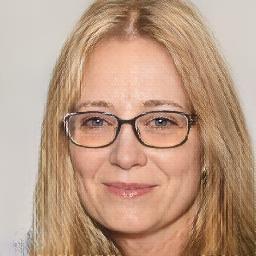} &
\includegraphics[width=0.12\linewidth]{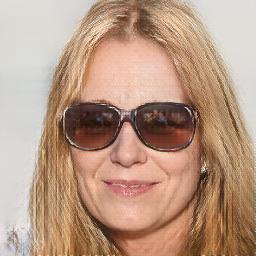} &
\includegraphics[width=0.12\linewidth]{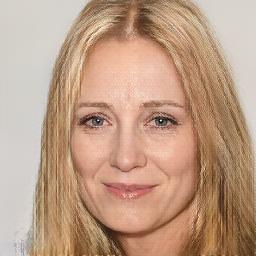} &
\includegraphics[width=0.12\linewidth]{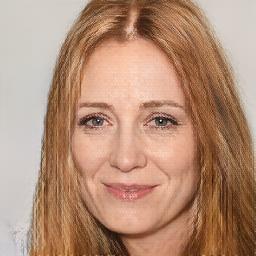} \\

\end{tabular}
\end{center}
\caption{Zero-shot manipulation of human faces. StyleGAN-based approaches mainly disentangle highly-localized visual concepts (e.g. glasses) while global concepts (e.g. gender) seem to be entangled with identity. Moreover, their manipulation requires manual calibration, leading to negligible changes (e.g. invisible glasses) or extreme edits (e.g. translation to asian does not preserve identity). LORD does not require calibration but struggles to disentangle attributes which are not perfectly uncorrelated (e.g. the gender attribute is ignored and remains entangled with hair color). Our method generates highly disentangled results without manual tuning.
}
\label{fig:ffhq}
\vspace{-1em}
\end{figure*}

\subsubsection{Results}
\begin{figure*}[t]
\begin{center}
\begin{tabular}{c@{\hskip1pt}c@{\hskip0pt}c@{\hskip0pt}c@{\hskip0pt}c@{\hskip0pt}c@{\hskip0pt}c@{\hskip0pt}c}

Input & \footnotesize Boerboel & \footnotesize Labradoodle & \footnotesize Chihuahua & \footnotesize {Bombay Cat} & \footnotesize Tiger & \footnotesize Lioness & \footnotesize{Arctic Fox} \\

\includegraphics[width=0.12\linewidth]{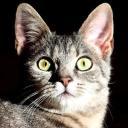} &
\includegraphics[width=0.12\linewidth]{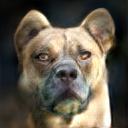} &
\includegraphics[width=0.12\linewidth]{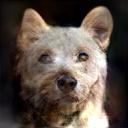} &
\includegraphics[width=0.12\linewidth]{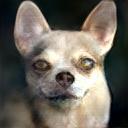} &
\includegraphics[width=0.12\linewidth]{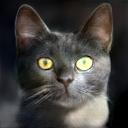} &
\includegraphics[width=0.12\linewidth]{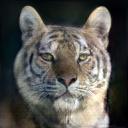} &
\includegraphics[width=0.12\linewidth]{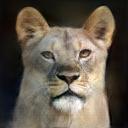} &
\includegraphics[width=0.12\linewidth]{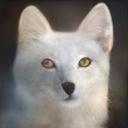} \\

\includegraphics[width=0.12\linewidth]{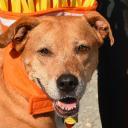} &
\includegraphics[width=0.12\linewidth]{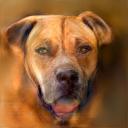} &
\includegraphics[width=0.12\linewidth]{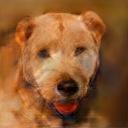} &
\includegraphics[width=0.12\linewidth]{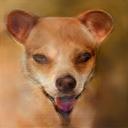} &
\includegraphics[width=0.12\linewidth]{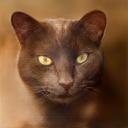} &
\includegraphics[width=0.12\linewidth]{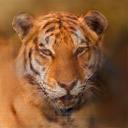} &
\includegraphics[width=0.12\linewidth]{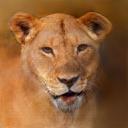} &
\includegraphics[width=0.12\linewidth]{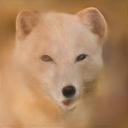} \\

\includegraphics[width=0.12\linewidth]{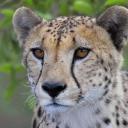} &
\includegraphics[width=0.12\linewidth]{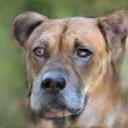} &
\includegraphics[width=0.12\linewidth]{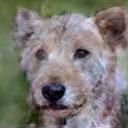} &
\includegraphics[width=0.12\linewidth]{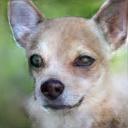} &
\includegraphics[width=0.12\linewidth]{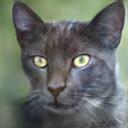} &
\includegraphics[width=0.12\linewidth]{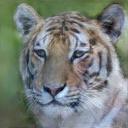} &
\includegraphics[width=0.12\linewidth]{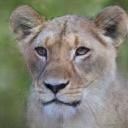} &
\includegraphics[width=0.12\linewidth]{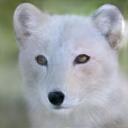} \\

\end{tabular}
\end{center}
\caption{Zero-shot translation results of animal species on AFHQ. The pose of the animal (which is never explicitly specified) is preserved reliably while synthesizing images of different species.}
\label{fig:afhq}
\end{figure*}

We compare our approach to two recent text-guided image manipulation techniques, TediGAN \cite{xia2021towards} and StyleCLIP \cite{patashnik2021styleclip}, both utilizing a pretrained StyleGAN generator and a pretrained CLIP network. As can be seen in Fig.~\ref{fig:ffhq}, despite their impressive image quality, methods that rely on a pretrained unconditional StyleGAN suffer from two critical drawbacks: (i) They disentangle mainly localized visual concepts (e.g. glasses and hair color), while the control of global concepts (e.g. gender) seems to be entangled with the face identity. (ii) The traversal in the latent space (the "strength" of the manipulation) is often tuned in a trial-and-error fashion for a given image and can not be easily calibrated across images, leading to unexpected results. For example, Fig.~\ref{fig:ffhq} shows two attempts to balance the manipulation strength in StyleCLIP ("-" denotes weaker manipulation and greater disentanglement threshold than "+"), although both seem suboptimal. Our manipulations are highly disentangled and obtained without any manual tuning. Fig.~\ref{fig:afhq} shows results of breed and species translation on AFHQ. The pose of the animal (which is never explicitly specified) is preserved reliably while synthesizing images of different species. Results of manipulating car types and colors are provided in Fig.~\ref{fig:cars}. More qualitative and quantitative comparisons are provided in the Appendix.

\subsection{Limitations}
As shown in our experiments, our principled disentanglement approach contributes significantly to the control and manipulation of real image attributes in the absence of full supervision. Nonetheless, two main limitations should be noted: (i) Although our method does not require manual annotation, it is trained for a \textit{fixed} set of attributes of interest, in contrast to methods as StyleCLIP which could adapt to a new visual concept at inference time. (ii) Unlike unconditional image generative models such as StyleGAN, reconstruction-based methods as ours struggle with synthesizing regions which exhibit large variability as hair-style. We believe that the trade-off between disentanglement and perceptual quality is an interesting research topic which is beyond the scope of this paper.

\begin{figure*}[t]
\begin{center}
\begin{tabular}{c@{\hskip1pt}c@{\hskip0pt}c@{\hskip0pt}c@{\hskip0pt}c@{\hskip0pt}c@{\hskip0pt}c@{\hskip0pt}c}

Input & Jeep & Sports & Family & Black & White & Red & Yellow \\

\includegraphics[width=0.12\linewidth]{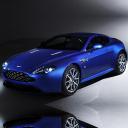} &
\includegraphics[width=0.12\linewidth]{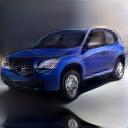} &
\includegraphics[width=0.12\linewidth]{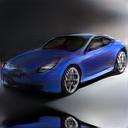} &
\includegraphics[width=0.12\linewidth]{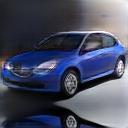} &
\includegraphics[width=0.12\linewidth]{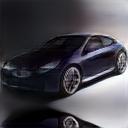} &
\includegraphics[width=0.12\linewidth]{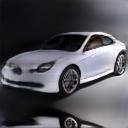} &
\includegraphics[width=0.12\linewidth]{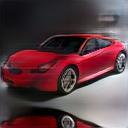} &
\includegraphics[width=0.12\linewidth]{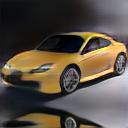} \\

\includegraphics[width=0.12\linewidth]{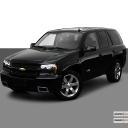} &
\includegraphics[width=0.12\linewidth]{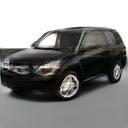} &
\includegraphics[width=0.12\linewidth]{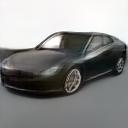} &
\includegraphics[width=0.12\linewidth]{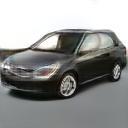} &
\includegraphics[width=0.12\linewidth]{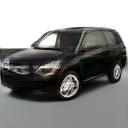} &
\includegraphics[width=0.12\linewidth]{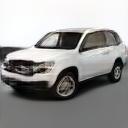} &
\includegraphics[width=0.12\linewidth]{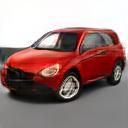} &
\includegraphics[width=0.12\linewidth]{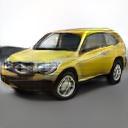} \\

\includegraphics[width=0.12\linewidth]{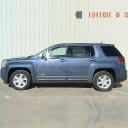} &
\includegraphics[width=0.12\linewidth]{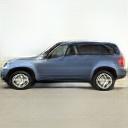} &
\includegraphics[width=0.12\linewidth]{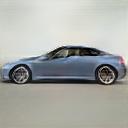} &
\includegraphics[width=0.12\linewidth]{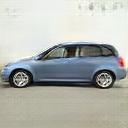} &
\includegraphics[width=0.12\linewidth]{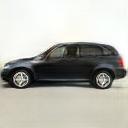} &
\includegraphics[width=0.12\linewidth]{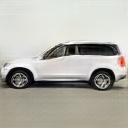} &
\includegraphics[width=0.12\linewidth]{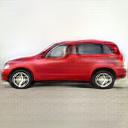} &
\includegraphics[width=0.12\linewidth]{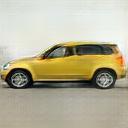} \\

\includegraphics[width=0.12\linewidth]{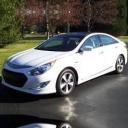} &
\includegraphics[width=0.12\linewidth]{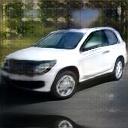} &
\includegraphics[width=0.12\linewidth]{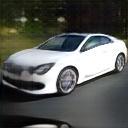} &
\includegraphics[width=0.12\linewidth]{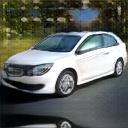} &
\includegraphics[width=0.12\linewidth]{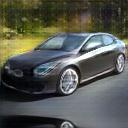} &
\includegraphics[width=0.12\linewidth]{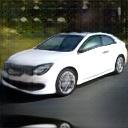} &
\includegraphics[width=0.12\linewidth]{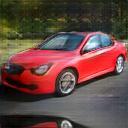} &
\includegraphics[width=0.12\linewidth]{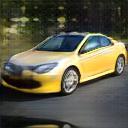} \\

\end{tabular}
\end{center}
\caption{Zero-shot translation results of car types and colors.}
\label{fig:cars}
\end{figure*}

\section{Conclusion}
\label{sec:conclusion}
We studied a disentanglement setting, in which few labels are given only for a limited subset of the underlying factors of variation, that better fits the modeling of real image distributions. We then proposed a novel disentanglement method which is shown to learn better representations than semi-supervised disentanglement methods on several synthetic benchmarks. Our robustness to partial labeling enables the use of zero-shot classifiers which can annotate (only) a partial set of visual concepts. Finally, we demonstrated better disentangled attribute manipulation of real images. We expect the core ideas proposed in this paper to carry over to other modalities and applications.

\section{Broader Impact}
\label{sec:impact}
Disentanglement of images in real life settings bears great potential societal impact.  On the positive side, better disentanglement may allow employing systems which are more invariant to protected attributes \cite{locatello2019fairness}. This may decrease the amount of discrimination which machine learning algorithms may exhibit when deployed on real life scenarios. On the negative side, along the possibility of malicious use of disentangled properties to discriminate intentionally; our work makes use of CLIP, a network pretrained on automatically collected images and labels. Such data are naturally prone to contain many biases. However, the technical contribution of our work can be easily adapted to future, less biased, pretrained networks.

The second aim of our work, namely, to produce disentangled zero shot image manipulations may present a broad impact as well. This is especially true when considering manipulation of human images. Disentangled image manipulation may assist synthetically creating more balanced datasets \cite{du2020fairness} in cases where the acquisition is extremely difficult (like rare disease). However, such methods may also be abused to create fake, misleading images \cite{wang2020cnn}. On top of that, the manipulation methods themselves may introduce new biases. For example, our method is reliant of a finite set of values, which is far from describing correctly many attributes. Therefore,
we stress that our method should be examined critically before use in tasks such as described here. We believe that dealing successfully with the raised issues in practice calls for future works, ranging from the technical sides, up to the regulatory and legislative ones. 

\paragraph{Acknowledgments} We thank the anonymous reviewers for their thoughtful review and constructive feedback that helped clarifying the contribution and framing of this paper. We are grateful to Prof. Shmuel Peleg for the suggestions regarding the presentation of our work. This work was partly supported by the Federmann Cyber Security Research Center in conjunction with the Israel National Cyber Directorate. Computational resources were kindly supplied by Oracle Cloud Services.

\bibliographystyle{plainnat}
\bibliography{neurips_2021}

\newpage
\clearpage

\appendix
\addcontentsline{toc}{section}{Appendix} 
\part{\Large Appendix - An Image is Worth More Than a Thousand Words: Towards Disentanglement in The Wild} 
\parttoc 

\section{Semi-Supervised Disentanglement with Residual Attributes}
\label{sec:appendix}

\subsection{Implementation Details}
\label{app:implementation_details}
Recall that our generative model gets as input the assignment of the attributes of interest together with the residual attributes and generates an image: 

\begin{equation}
x = G(\tilde{f}^1, ..., \tilde{f}^k, r)
\end{equation}

where $\tilde{f}^j$ is the output probability vector (over $m^j$ different classes) of the classifier $C^j$. 
Generally, the embedding dimension of an attribute of interest $j$ within the generator is $d^j$. The embedding of each attribute of interest is obtained by the following projection: $\tilde{f}^j \cdot P^j$ where $P^j \in \mathbb{R}^{m^j \times d^j}$. The parameters of $P^j$ are optimized together with the rest of the parameters of the generator $G$. All the representations of the attributes of interest are concatenated along with the representation of the residual attributes before being fed into the generator $G$.

\paragraph{Architecture for Synthetic Experiments} 
To be inline with the disentanglement literature, we set $d^j = 1$ in the synthetic experiments, i.e. each attribute of interest should be represented in a single latent dimension. The entire latent code is therefore composed of $d = 10$ dimensions, $k$ of which are devoted for $k$ attributes of interest and $10 - k$ are for the residual attributes.
The generator $G$ is an instance of the architecture of the betaVAE decoder, while each of the classifiers $C^j$ and the residual encoder $E_r$ is of the betaVAE encoder form. Detailed architectures are provided in  Tab.~\ref{tab:generator_betavae} and Tab.~\ref{tab:encoder_betavae} for completeness.

\paragraph{Architecture for Experiments on Real Images}
In order to scale to real images in high resolution, we replace the betaVAE architecture with the generator architecture of StyleGAN2, with two modifications; (i) The latent code is adapted to our formulation and forms a concatenation of the representations of the attributes of interest and the residual attributes. (ii) We do not apply any noise injection, unlike the unconditional training of StyleGAN2. Note that in these experiments we train an additional adversarial discriminator to increase the perceptual quality of the synthesized images. A brief summary of the architectures is presented in Tab.~\ref{tab:generator_stylegan2} and \ref{tab:discriminator_stylegan2} for completeness. The architecture of the feed-forward residual encoder trained in the second stage is influenced by StarGAN-v2 \cite{choi2020stargan} and presented in Tab.~\ref{tab:encoder_stargan2}.

\paragraph{Optimization} All the modules of our disentanglement model are trained from scratch, including the generator $G$, classifiers $C^1, ..., C^k$ and a residual latent code $r_i$ per image $x_i$. While in the synthetic experiments each of the $k$ attributes of interest is embedded into a single dimension, we set the dimension of each attribute to $8$ and of the residuals to $256$ in the experiments on real images. We set the learning rate of the latent codes to $0.01$, of the generator to $0.001$ and of the attribute classifiers to $0.0001$. The learning rate of the additional discriminator (only trained for real images) is set to 0.0001. The practice of using a higher learning rate for the latent codes is motivated by the fact that the latent codes (one per image) are updated only once in an entire epoch, while the parameters of the other modules are updated in each mini-batch. For each mini-batch, we update the parameters of the models and the relevant latent codes with a single gradient step each. The different loss weights are set to $\lambda_{cls} = 0.001, \lambda_{ent} = 0.001, \lambda_{res} = 0.0001$. In order to stabilize the training, we sample both supervised and unsupervised samples in each mini-batch i.e. half of the images in each mini-batch are labeled with at least one true attribute of interest.

\begin{table}
  \caption{Attribute splits for the synthetic benchmarks.}
  \label{tab:synthetic_attribute_splits}
  \centering
  \begin{tabular}{lll}
    \toprule
    Dataset & Attributes of Interest &	Residual Attributes \\
    \midrule
    Shapes3D & floor color, wall color, object color & scale, shape, azimuth \\
    Cars3D & elevation, azimuth & object \\
    dSprites & scale, x, y & orientation, shape \\
    SmallNORB & elevation, azimuth, lighting & category, instance \\

    \bottomrule
  \end{tabular}
\end{table}

\subsection{Evaluation Protocol}
We assess the learned representations of the attributes of interest using DCI \cite{eastwood2018dci} which measures three properties: (i) \textit{Disentanglement} - the degree to which each variable (or dimension) captures at most one generative factor. (ii) \textit{Completeness} - the degree to which each underlying factor is captured by a single variable (or dimension). (iii) \textit{Informativeness} - the total amount of information that a representation captures about the underlying
factors of variation. Tab.~\ref{tab:synthetic_quantitative} summarizes the quantitative evaluation of our method and the baselines on the synthetic benchmarks using DCI and two other disentanglement metrics: SAP \cite{kumar2017sapscore} and MIG \cite{chen2018tcvae}.

Regarding the fully unlabeled residual attributes, we only require the learned representation to be informative of the residual attributes and disentangled from the attributes of interest. As we cannot expect the codes here to be disentangled from one another, we cannot use the standard disentanglement metrics (e.g. DCI). For evaluating these criteria, we train a set of linear classifiers, each of which attempts to predict a single attribute given the residual representations (using the available true labels). The results and comparisons can be found in Sec.~\ref{app:additional_results}.

The specific attribute splits used in our synthetic experiments are provided in Tab.~\ref{tab:synthetic_attribute_splits}.

\subsection{Baseline Models}
\subsubsection{Synthetic Experiments}

\textbf{Semi-Supervised betaVAE (\citet{locatello2020fewlabels}):} We compare with the original implementation\footnote{https://github.com/google-research/disentanglement\_lib} of the semi-supervised variant of betaVAE provided by \citet{locatello2020fewlabels}. Note that we  consider the \textit{permuted-labels} configuration in which the attribute values are not assumed to exhibit a semantic order. While the order of the values can be exploited as an inductive bias to disentanglement, many attributes in the real world (e.g. human gender or animal specie) are not ordered in a meaningful way. 

\textbf{LORD (\citet{gabbay2020lord}):}
As this method assumes full-supervision on the attributes of interest, we make an effort to adapt to the proposed setting and regularize the latent codes of the unlabeled images with activation decay penalty and Gaussian noise. We do it in a similar way to the regularization applied to the residual latent codes. While other forms of regularization might be considered, we believe it is the most trivial extension of LORD to support partially-labeled attributes. 

\subsubsection{Real Images Experiments}

\textbf{LORD (\citet{gabbay2020lord}):} We use the same method as in the synthetic experiments, but adapt it to real images. Similarly to the adaptation applied for our method, we replace the betaVAE architecture with StyleGAN2, add adversarial discriminator, and increase the latent codes dimensions. The limited supervision here is supplied by CLIP \cite{radford2021clip} as in our method.


\textbf{StyleCLIP (\citet{patashnik2021styleclip}):} We used the official repository\footnote{https://github.com/orpatashnik/StyleCLIP} of the authors. The parameters were optimized to obtain both visually pleasing and disentangled results. Two configurations were chosen, one for the "StyleCLIP+" comparison ($\alpha = 4.1$), and one for "StyleCLIP-" ($\alpha = 2.1$). We always used $\beta = 0.1$. 

\textbf{TediGAN (\citet{xia2021towards}):} To optimize the trade-off between visually pleasing results and applying the desired manipulation, the value of $"clip\_loss" = 5.0$ was chosen. Other parameters were given their default value in the official supplied code\footnote{https://github.com/IIGROUP/TediGAN}.

We note that to obtain the results presented in StyleCLIP and TediGAN papers, the authors adjusted these parameters per image. While this might be a reasonable practice for an artist utilizing these methods, it is not part of our setting.

\subsubsection{Relation to LORD \cite{gabbay2020lord}}
Recall that we aim to achieve disentanglement in the absence of full supervision on the attributes of interest (i.e. the supervised class in \cite{gabbay2020lord}). We stress and show that methods as LORD \cite{gabbay2020lord} that indeed aim to disentangle the attributes of interest from a unified set of residual attributes, only work when full supervision is available on the attributes of interest and struggle when only partial labels are provided (see Tab.~\ref{tab:synthetic_quantitative} and Fig.~\ref{fig:ffhq_additional1},\ref{fig:ffhq_additional2},\ref{fig:ffhq_additional3}). More specifically, our method can be seen as an extension of LORD \cite{gabbay2020lord} for cases where the attributes of interest are observed only in a very few samples. Our method can therefore also leverage off-the-shelf zero-shot image classifiers such as CLIP, in order to be applied without the need to manually annotate even a small set of images. The few labels obtained with CLIP can not be used effectively by LORD \cite{gabbay2020lord}, as demonstrated in our experiments.

From a technical perspective, there are two fundamental differences between our method and LORD \cite{gabbay2020lord}: (i) LORD is a fully latent-based model i.e. no classifiers are trained with the generator in the first stage. The latent codes of the attributes of interest are optimized directly (and shared between all instances with the same label). Here we provide a hybrid latent-amortized approach where attribute codes are learned in a latent fashion, similarly to LORD, but they are weighted using the probabilities emitted by an amortized classifier. (ii) We introduce an additional term $\mathcal{L}_{ent}$ which enables our method to perform well when very limited supervision exists for the attributes of interest.

\subsection{Additional Results}
\label{app:additional_results}

\subsubsection{Synthetic Experiments}
\label{app:add_synth_exp}

\paragraph{Disentanglement of the Residual Code} We report the accuracy of linear classifiers in predicting the values of the different attributes from the residual code in Tab.~\ref{tab:synthetic_quantitative_residual}. Ideally, the residual code should contain all the information about the residual attributes, and no information about the attributes of interest. Therefore, for each dataset we expect the attributes of interest (first row of each dataset, colored in red) to be predicted with low accuracy. The residual attributes (second row, in green) are the ones that \textit{should} be encoded by the residual code. 

We see that our method almost always provides better disentanglement (worse prediction) between the residual code and the attributes of interest. Although the residual code in the method by \citet{locatello2020fewlabels} sometimes provide better predictions regarding the residual attributes, it comes at the expense of containing a lot information regarding the attributes of interest. Keeping in mind that our goal is to disentangle the attributes of interest between themselves, and from the residual code, we conclude the our suggested method performs better on this metric as well.

\paragraph{Regularization Terms} We provide an ablation study of the different terms in Eq.~\ref{eq:loss_disentanglement}. We first show in Tab.~\ref{tab:ent_abl} that without the entropy penalty $\mathcal{L}_{ent}$ we obtain inferior disentanglement of the attributes of interest. For evluating the importance of the residual codes regularization $\mathcal{L}_{res}$ we measure the accuracy of classifying the attributes of interest from the residual representations using logistic regression. The results in Tab.~\ref{tab:ent_res} highlight the contribution of this term.

\paragraph{Pseudo-labels} We show in Tab.~\ref{tab:ablation_full} a full version of the average ablation table of attribute classification supplied on the main text (Tab.~\ref{tab:ablation_quantitative}). The results suggest that the attribute classification accuracy is improved by our method, compared to using the same architecture as a classifier trained only on the labeled samples.  

\begin{table}[t]
  \caption{Ablation for $\mathcal{L}_{ent}$ using 1000 [or 100] labels per attribute of interest.}
  \label{tab:ent_abl}
  \centering
  \begin{tabular}{lllllll}
        \toprule
  & & D & C & I & SAP & MIG \\
    \midrule
\multirow{2}{*}{Shapes3D}	 &	 Ours w/o  $\mathcal{L}_{ent}$	&	0.99	[0.99]	&	0.98	[0.98] 	&	0.98	[0.98] 	&	0.28	[0.26] 	&	0.94	[0.91]   	\\
& Ours	&	\textbf{1.00}	\textbf{[1.00]}	&	\textbf{1.00}	\textbf{[1.00]}	&	\textbf{1.00}	\textbf{[1.00]}	&	\textbf{0.30}	\textbf{[0.30]} 	&	\textbf{1.00}	\textbf{[0.96]}\\
    \midrule
\multirow{2}{*}{Cars3D}	 &	 Ours w/o  $\mathcal{L}_{ent}$	&	0.74	[0.39]	&	0.74	[0.40]	&	0.71	[0.43]	&	0.22	[0.11]	&	0.57	[0.34]	\\
& Ours	&	\textbf{0.80}	\textbf{[0.40]} &	\textbf{0.80}	\textbf{[0.41]} &	\textbf{0.78}	\textbf{[0.56]} &	\textbf{0.33}	\textbf{[0.15]} &	\textbf{0.61}	\textbf{[0.35]}\\
    \bottomrule
  \end{tabular}
\end{table}

\begin{table}[t]
  \caption{Ablation for $\mathcal{L}_{res}$ using 1000 labels per attribute of interest (lower is better).}
  \label{tab:ent_res}
  \centering
  \begin{tabular}{lllll}
        \toprule
  & & floor color & wall color & object color \\
    \midrule
\multirow{3}{*}{Shapes3D}   &	 Ours w/o 	 $\mathcal{L}_{res}$	&	0.23	&	0.28	&	0.18	\\
& Ours	&	\textbf{0.11} &	\textbf{0.12} &	\textbf{0.14} \\
& Random Chance (optimal)	&	 0.10	&	0.10	&	0.10	\\
    \bottomrule
  \end{tabular}
\end{table}

\subsubsection{Real Images Experiments}

\paragraph{Quantitative Evaluation} The quantitative evaluation of disentanglement in real images is challenging as no ground truth annotations are available for all the attributes and the attributes are not completely independent. For evaluation purposes, the paper includes quantitative metrics on synthetic benchmarks and many qualitative comparisons on real images. We further consider quantitative metrics for evaluation of our method on real images of human faces. We assess the performance by Attribute-Dependency (AD) (proposed in \cite{wu2021stylespace}): we measure the degree to which manipulation of a certain attribute induces changes in other attributes, as measured by classifiers for these attributes. We rely on 40 pretrained classifiers for attributes in CelebA, in order to cope with real images, where the exact factors of variation are not observed. Intuitively, disentangled manipulations should induce smaller changes in other attributes (lower AD is better). In addition, we report the \textit{manipulation strength} for each attribute, as measured by the normalized change to the logit of the classifier of the target attribute. Note that the manipulation strength can be negative in cases where the manipulation causes an opposite effect to the attribute. 

\begin{figure*}[t]
\begin{center}
\begin{tabular}{@{\hskip0pt}c@{\hskip2pt}c@{\hskip3pt}c@{\hskip0pt}c@{\hskip0pt}c@{\hskip0pt}c@{\hskip0pt}c@{\hskip0pt}c@{\hskip0pt}c}

& Input & Kid & Asian & Gender & Glasses & Shades & Beard & Red hair \\

\begin{turn}{90} ~ TediGAN \end{turn} &
\includegraphics[width=0.12\linewidth]{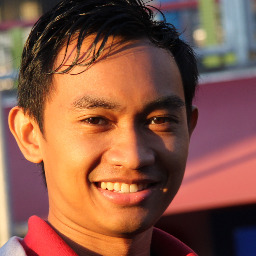} &
\includegraphics[width=0.12\linewidth]{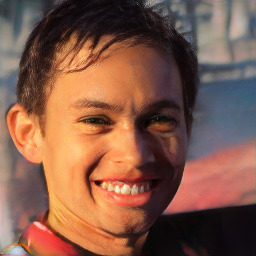} &
\includegraphics[width=0.12\linewidth]{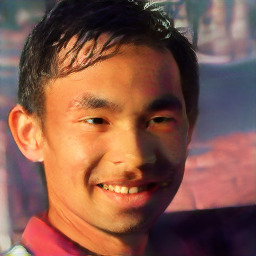} &
\includegraphics[width=0.12\linewidth]{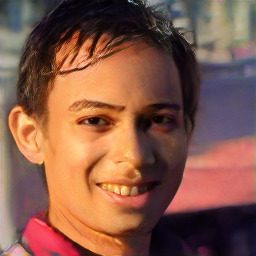} &
\includegraphics[width=0.12\linewidth]{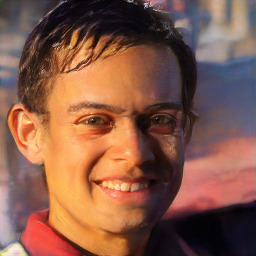} &
\includegraphics[width=0.12\linewidth]{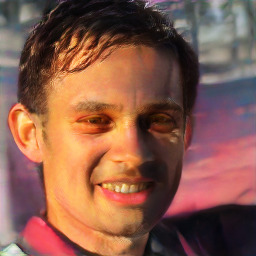} &
\includegraphics[width=0.12\linewidth]{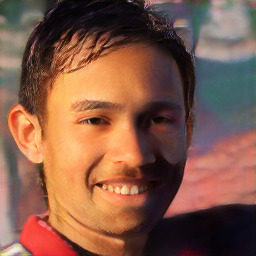} &
\includegraphics[width=0.12\linewidth]{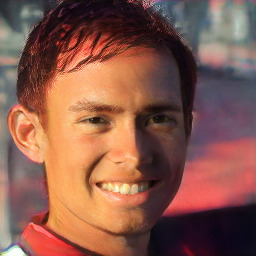} \\

\begin{turn}{90} ~ \footnotesize StyleCLIP- \end{turn} &
\includegraphics[width=0.12\linewidth]{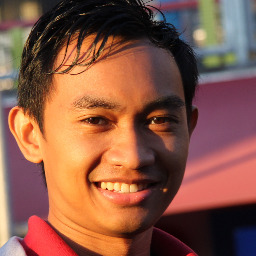} &
\includegraphics[width=0.12\linewidth]{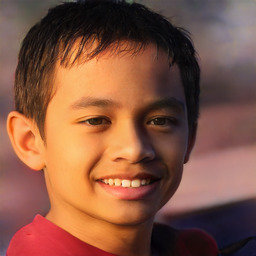} &
\includegraphics[width=0.12\linewidth]{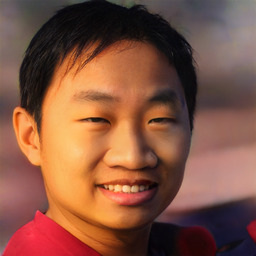} &
\includegraphics[width=0.12\linewidth]{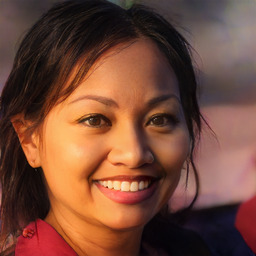} &
\includegraphics[width=0.12\linewidth]{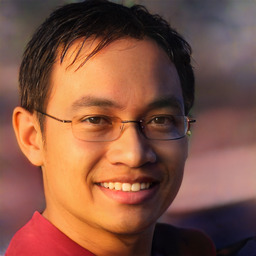} &
\includegraphics[width=0.12\linewidth]{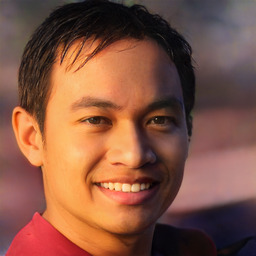} &
\includegraphics[width=0.12\linewidth]{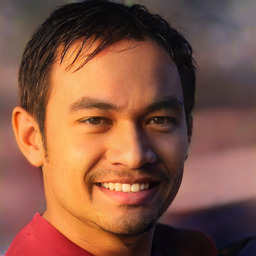} &
\includegraphics[width=0.12\linewidth]{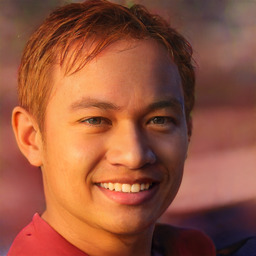} \\

\begin{turn}{90} \footnotesize ~ StyleCLIP+ \end{turn} &
\includegraphics[width=0.12\linewidth]{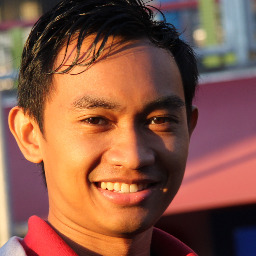} &
\includegraphics[width=0.12\linewidth]{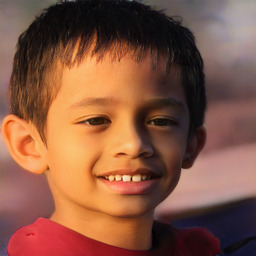} &
\includegraphics[width=0.12\linewidth]{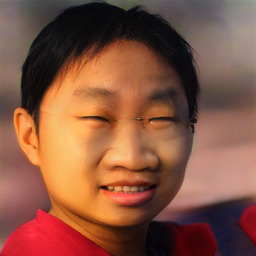} &
\includegraphics[width=0.12\linewidth]{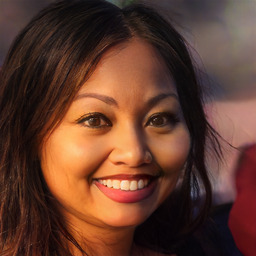} &
\includegraphics[width=0.12\linewidth]{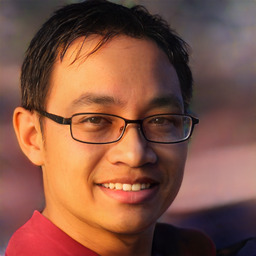} &
\includegraphics[width=0.12\linewidth]{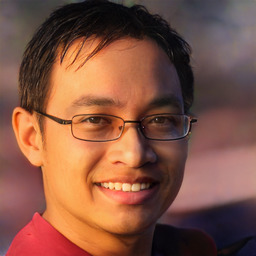} &
\includegraphics[width=0.12\linewidth]{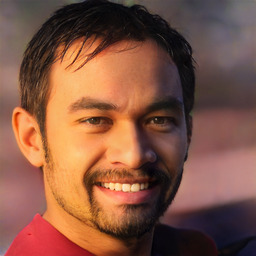} &
\includegraphics[width=0.12\linewidth]{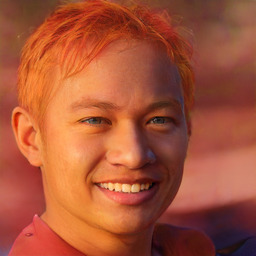} \\

\begin{turn}{90} ~~~ LORD \end{turn} &
\includegraphics[width=0.12\linewidth]{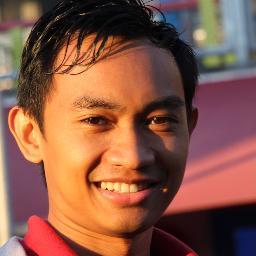} &
\includegraphics[width=0.12\linewidth]{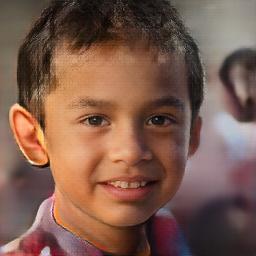} &
\includegraphics[width=0.12\linewidth]{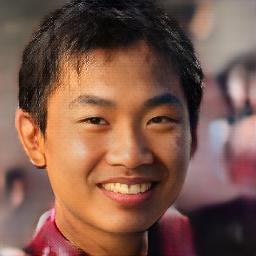} &
\includegraphics[width=0.12\linewidth]{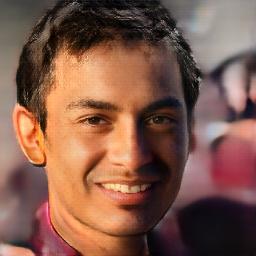} &
\includegraphics[width=0.12\linewidth]{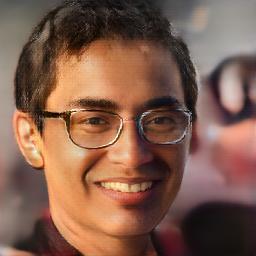} &
\includegraphics[width=0.12\linewidth]{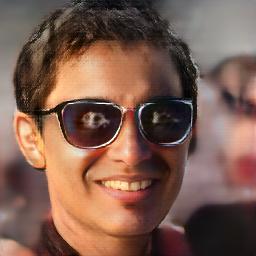} &
\includegraphics[width=0.12\linewidth]{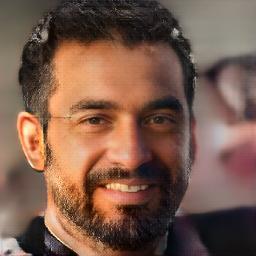} &
\includegraphics[width=0.12\linewidth]{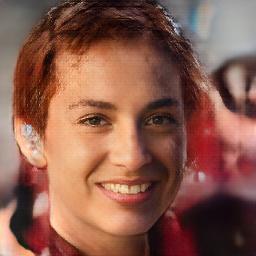} \\

\begin{turn}{90} ~~~~ \textbf{Ours} \end{turn} &
\includegraphics[width=0.12\linewidth]{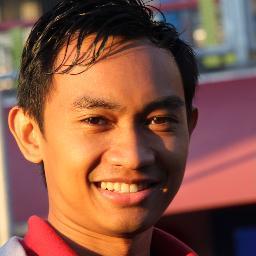} &
\includegraphics[width=0.12\linewidth]{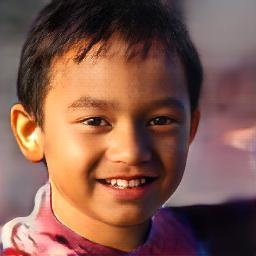} &
\includegraphics[width=0.12\linewidth]{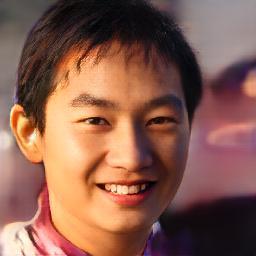} &
\includegraphics[width=0.12\linewidth]{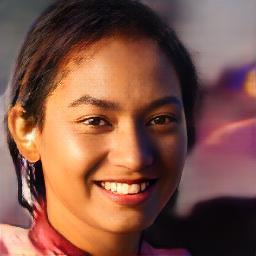} &
\includegraphics[width=0.12\linewidth]{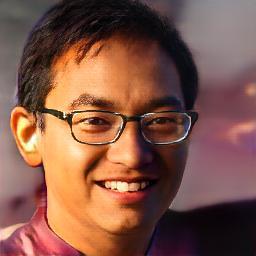} &
\includegraphics[width=0.12\linewidth]{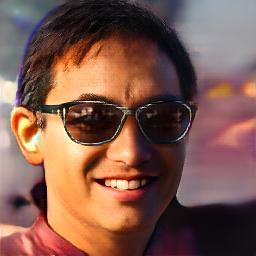} &
\includegraphics[width=0.12\linewidth]{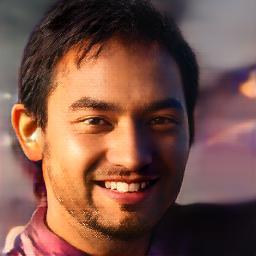} &
\includegraphics[width=0.12\linewidth]{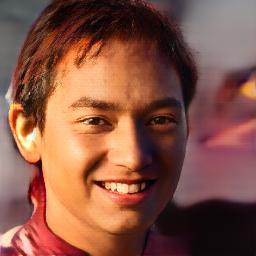} \\

\begin{turn}{90} ~ TediGAN \end{turn} &
\includegraphics[width=0.12\linewidth]{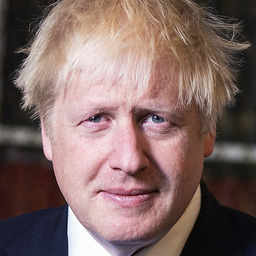} &
\includegraphics[width=0.12\linewidth]{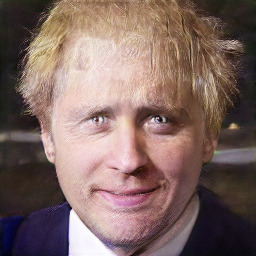} &
\includegraphics[width=0.12\linewidth]{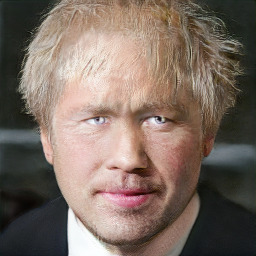} &
\includegraphics[width=0.12\linewidth]{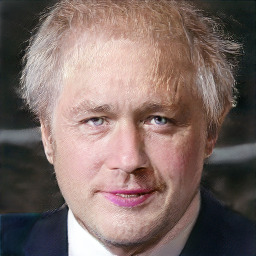} &
\includegraphics[width=0.12\linewidth]{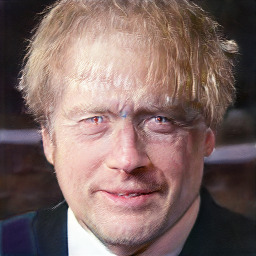} &
\includegraphics[width=0.12\linewidth]{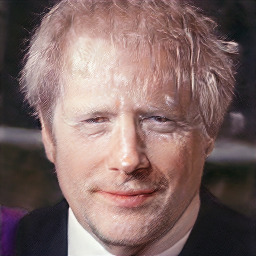} &
\includegraphics[width=0.12\linewidth]{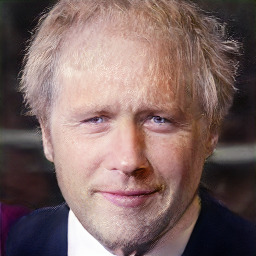} &
\includegraphics[width=0.12\linewidth]{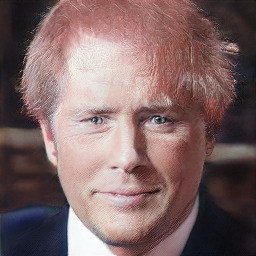} \\

\begin{turn}{90} ~ \footnotesize StyleCLIP- \end{turn} &
\includegraphics[width=0.12\linewidth]{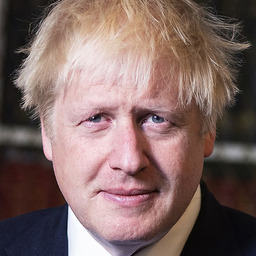} &
\includegraphics[width=0.12\linewidth]{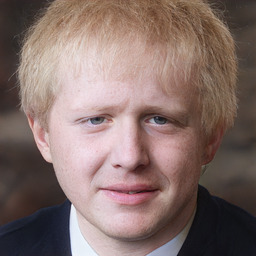} &
\includegraphics[width=0.12\linewidth]{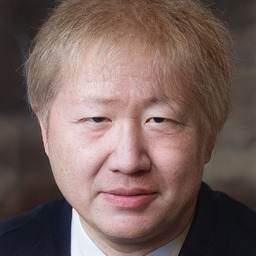} &
\includegraphics[width=0.12\linewidth]{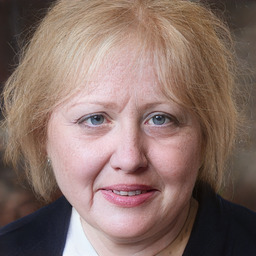} &
\includegraphics[width=0.12\linewidth]{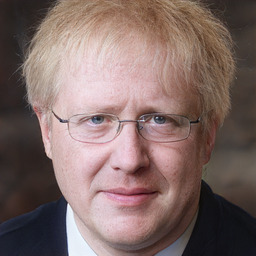} &
\includegraphics[width=0.12\linewidth]{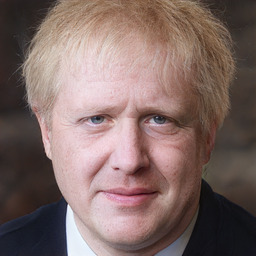} &
\includegraphics[width=0.12\linewidth]{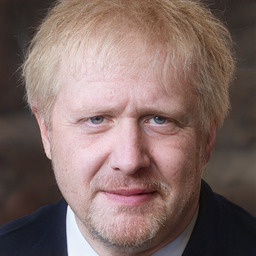} &
\includegraphics[width=0.12\linewidth]{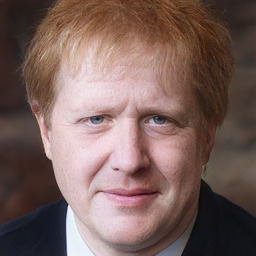} \\

\begin{turn}{90} \footnotesize ~ StyleCLIP+ \end{turn} &
\includegraphics[width=0.12\linewidth]{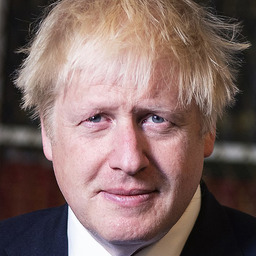} &
\includegraphics[width=0.12\linewidth]{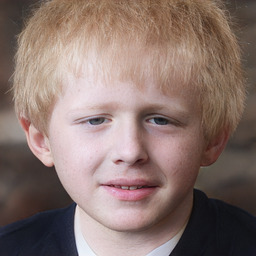} &
\includegraphics[width=0.12\linewidth]{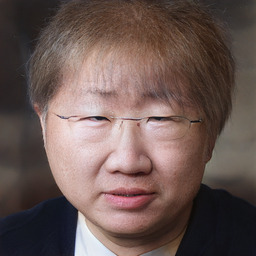} &
\includegraphics[width=0.12\linewidth]{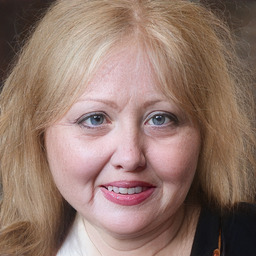} &
\includegraphics[width=0.12\linewidth]{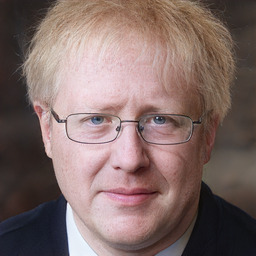} &
\includegraphics[width=0.12\linewidth]{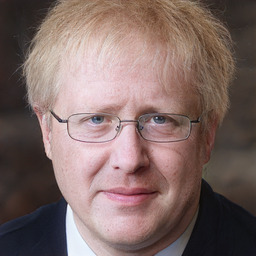} &
\includegraphics[width=0.12\linewidth]{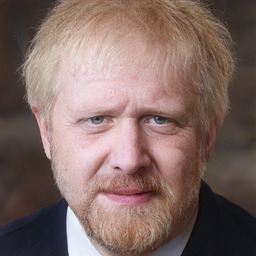} &
\includegraphics[width=0.12\linewidth]{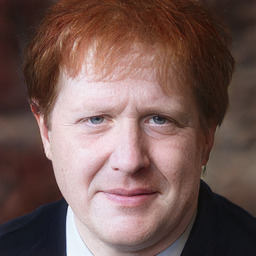} \\

\begin{turn}{90} ~~~ LORD \end{turn} &
\includegraphics[width=0.12\linewidth]{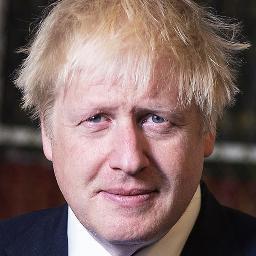} &
\includegraphics[width=0.12\linewidth]{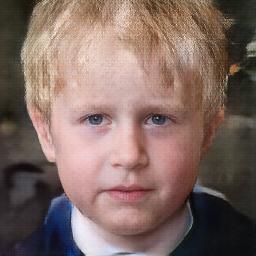} &
\includegraphics[width=0.12\linewidth]{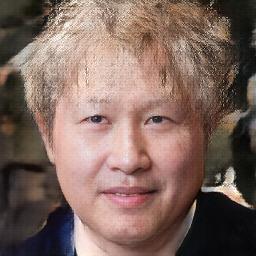} &
\includegraphics[width=0.12\linewidth]{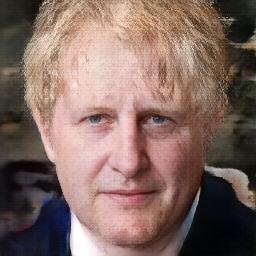} &
\includegraphics[width=0.12\linewidth]{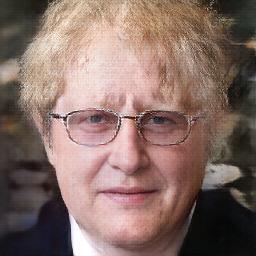} &
\includegraphics[width=0.12\linewidth]{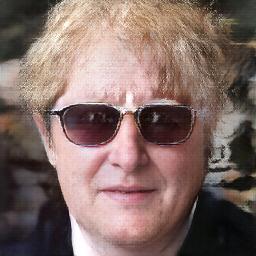} &
\includegraphics[width=0.12\linewidth]{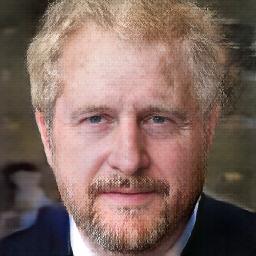} &
\includegraphics[width=0.12\linewidth]{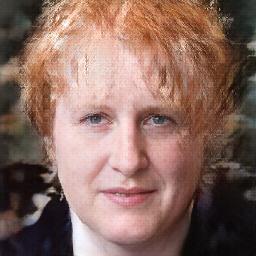} \\

\begin{turn}{90} ~~~~ \textbf{Ours} \end{turn} &
\includegraphics[width=0.12\linewidth]{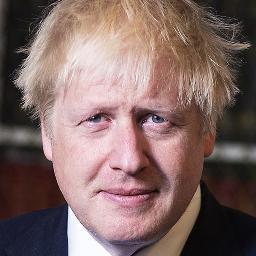} &
\includegraphics[width=0.12\linewidth]{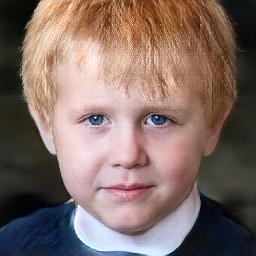} &
\includegraphics[width=0.12\linewidth]{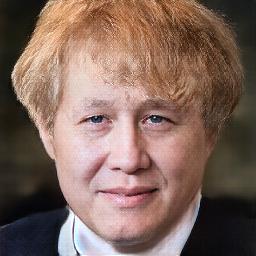} &
\includegraphics[width=0.12\linewidth]{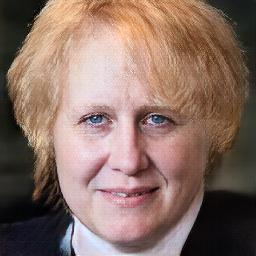} &
\includegraphics[width=0.12\linewidth]{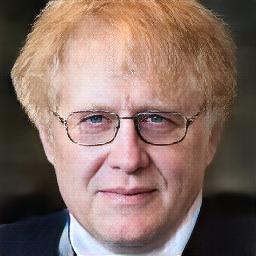} &
\includegraphics[width=0.12\linewidth]{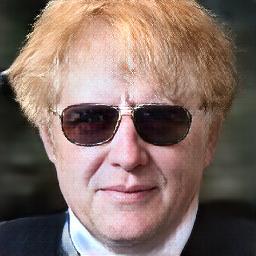} &
\includegraphics[width=0.12\linewidth]{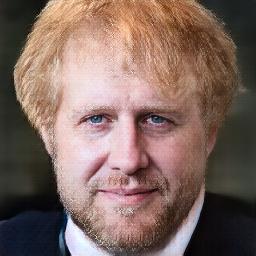} &
\includegraphics[width=0.12\linewidth]{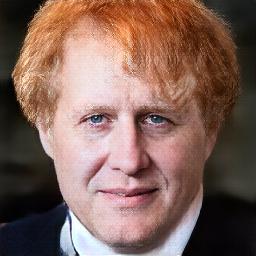} \\

\end{tabular}
\end{center}
\caption{Zero-shot manipulation of human faces. StyleGAN-based approaches  (TediGAN and StyleCLIP) mainly disentangle highly-localized visual concepts (e.g. beard) while global concepts (e.g. gender) seem to be entangled with identity and expression. Moreover, their manipulation requires manual calibration, leading to negligible changes (e.g. invisible glasses) or extreme edits (e.g. translation to asian does not preserve identity). LORD does not require calibration but struggles to disentangle attributes which are not perfectly uncorrelated (e.g. the gender attribute is ignored and remains entangled with beard and hair color). Our method generates highly disentangled results without manual tuning.}
\label{fig:ffhq_additional1}
\end{figure*}

\begin{figure*}[t]
\begin{center}
\begin{tabular}{@{\hskip0pt}c@{\hskip2pt}c@{\hskip3pt}c@{\hskip0pt}c@{\hskip0pt}c@{\hskip0pt}c@{\hskip0pt}c@{\hskip0pt}c@{\hskip0pt}c}

& Input & Kid & Asian & Gender & Glasses & Shades & Beard & Red hair \\

\begin{turn}{90} ~ TediGAN \end{turn} &
\includegraphics[width=0.12\linewidth]{figures/ffhq_additional/tedigan/e/input.jpg} &
\includegraphics[width=0.12\linewidth]{figures/ffhq_additional/tedigan/e/kid.jpg} &
\includegraphics[width=0.12\linewidth]{figures/ffhq_additional/tedigan/e/asian.jpg} &
\includegraphics[width=0.12\linewidth]{figures/ffhq_additional/tedigan/e/gender.jpg} &
\includegraphics[width=0.12\linewidth]{figures/ffhq_additional/tedigan/e/eyeglasses.jpg} &
\includegraphics[width=0.12\linewidth]{figures/ffhq_additional/tedigan/e/sunglasses.jpg} &
\includegraphics[width=0.12\linewidth]{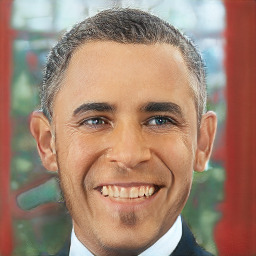} &
\includegraphics[width=0.12\linewidth]{figures/ffhq_additional/tedigan/e/red_hair.jpg} \\

\begin{turn}{90} ~ \footnotesize StyleCLIP- \end{turn} &
\includegraphics[width=0.12\linewidth]{figures/ffhq_additional/styleclip/alpha_2.1/e/input.jpg} &
\includegraphics[width=0.12\linewidth]{figures/ffhq_additional/styleclip/alpha_2.1/e/kid.jpg} &
\includegraphics[width=0.12\linewidth]{figures/ffhq_additional/styleclip/alpha_2.1/e/asian.jpg} &
\includegraphics[width=0.12\linewidth]{figures/ffhq_additional/styleclip/alpha_2.1/e/gender.jpg} &
\includegraphics[width=0.12\linewidth]{figures/ffhq_additional/styleclip/alpha_2.1/e/eyeglasses.jpg} &
\includegraphics[width=0.12\linewidth]{figures/ffhq_additional/styleclip/alpha_2.1/e/sunglasses.jpg} &
\includegraphics[width=0.12\linewidth]{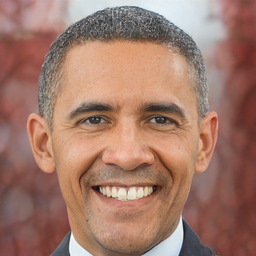} &
\includegraphics[width=0.12\linewidth]{figures/ffhq_additional/styleclip/alpha_2.1/e/red_hair.jpg} \\

\begin{turn}{90} \footnotesize ~ StyleCLIP+ \end{turn} &
\includegraphics[width=0.12\linewidth]{figures/ffhq_additional/styleclip/alpha_4.1/e/input.jpg} &
\includegraphics[width=0.12\linewidth]{figures/ffhq_additional/styleclip/alpha_4.1/e/kid.jpg} &
\includegraphics[width=0.12\linewidth]{figures/ffhq_additional/styleclip/alpha_4.1/e/asian.jpg} &
\includegraphics[width=0.12\linewidth]{figures/ffhq_additional/styleclip/alpha_4.1/e/gender.jpg} &
\includegraphics[width=0.12\linewidth]{figures/ffhq_additional/styleclip/alpha_4.1/e/eyeglasses.jpg} &
\includegraphics[width=0.12\linewidth]{figures/ffhq_additional/styleclip/alpha_4.1/e/sunglasses.jpg} &
\includegraphics[width=0.12\linewidth]{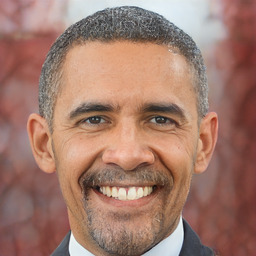} &
\includegraphics[width=0.12\linewidth]{figures/ffhq_additional/styleclip/alpha_4.1/e/red_hair.jpg} \\

\begin{turn}{90} ~~~ LORD \end{turn} &
\includegraphics[width=0.12\linewidth]{figures/ffhq_additional/lord/e/input.jpg} &
\includegraphics[width=0.12\linewidth]{figures/ffhq_additional/lord/e/kid.jpg} &
\includegraphics[width=0.12\linewidth]{figures/ffhq_additional/lord/e/asian.jpg} &
\includegraphics[width=0.12\linewidth]{figures/ffhq_additional/lord/e/female.jpg} &
\includegraphics[width=0.12\linewidth]{figures/ffhq_additional/lord/e/eyeglasses.jpg} &
\includegraphics[width=0.12\linewidth]{figures/ffhq_additional/lord/e/sunglasses.jpg} &
\includegraphics[width=0.12\linewidth]{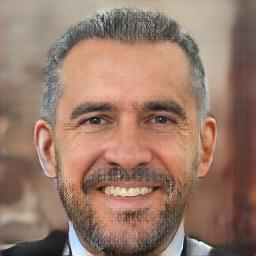} &
\includegraphics[width=0.12\linewidth]{figures/ffhq_additional/lord/e/red_hair.jpg} \\

\begin{turn}{90} ~~~~ \textbf{Ours} \end{turn} &
\includegraphics[width=0.12\linewidth]{figures/ffhq_additional/ours/e/input.jpg} &
\includegraphics[width=0.12\linewidth]{figures/ffhq_additional/ours/e/kid.jpg} &
\includegraphics[width=0.12\linewidth]{figures/ffhq_additional/ours/e/asian.jpg} &
\includegraphics[width=0.12\linewidth]{figures/ffhq_additional/ours/e/female.jpg} &
\includegraphics[width=0.12\linewidth]{figures/ffhq_additional/ours/e/eyeglasses.jpg} &
\includegraphics[width=0.12\linewidth]{figures/ffhq_additional/ours/e/sunglasses.jpg} &
\includegraphics[width=0.12\linewidth]{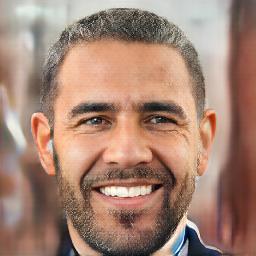} &
\includegraphics[width=0.12\linewidth]{figures/ffhq_additional/ours/e/red_hair.jpg} \\

\begin{turn}{90} ~ TediGAN \end{turn} &
\includegraphics[width=0.12\linewidth]{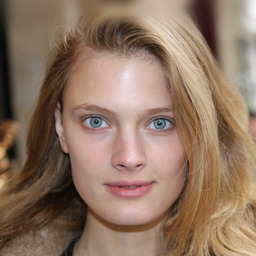} &
\includegraphics[width=0.12\linewidth]{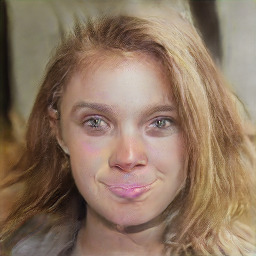} &
\includegraphics[width=0.12\linewidth]{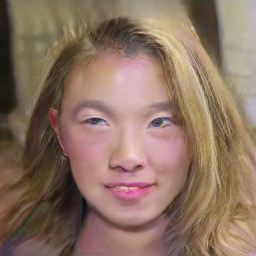} &
\includegraphics[width=0.12\linewidth]{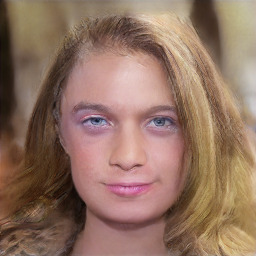} &
\includegraphics[width=0.12\linewidth]{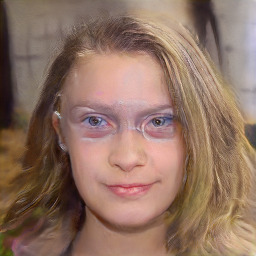} &
\includegraphics[width=0.12\linewidth]{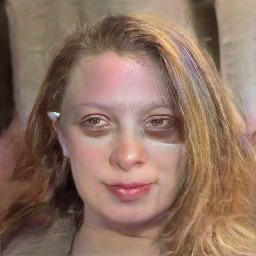} &
\includegraphics[width=0.12\linewidth]{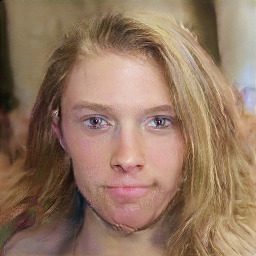} &
\includegraphics[width=0.12\linewidth]{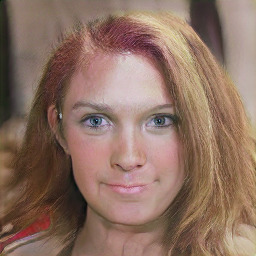} \\

\begin{turn}{90} ~ \footnotesize StyleCLIP- \end{turn} &
\includegraphics[width=0.12\linewidth]{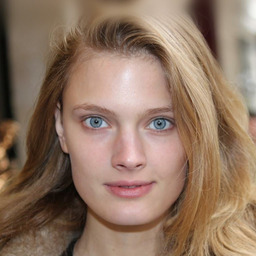} &
\includegraphics[width=0.12\linewidth]{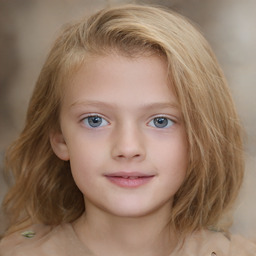} &
\includegraphics[width=0.12\linewidth]{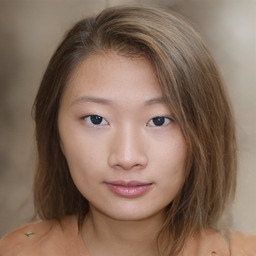} &
\includegraphics[width=0.12\linewidth]{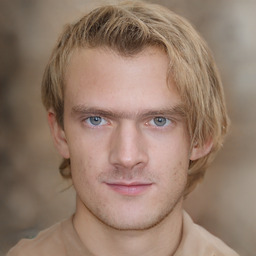} &
\includegraphics[width=0.12\linewidth]{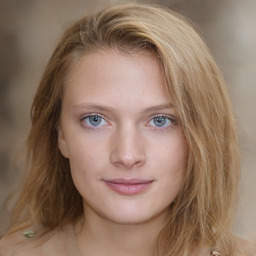} &
\includegraphics[width=0.12\linewidth]{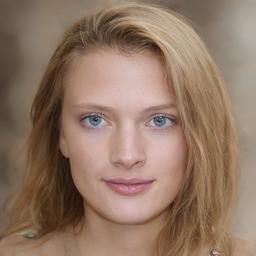} &
\includegraphics[width=0.12\linewidth]{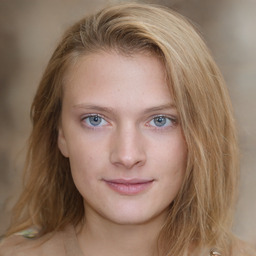} &
\includegraphics[width=0.12\linewidth]{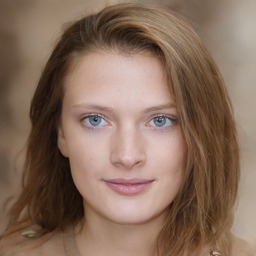} \\

\begin{turn}{90} \footnotesize ~ StyleCLIP+ \end{turn} &
\includegraphics[width=0.12\linewidth]{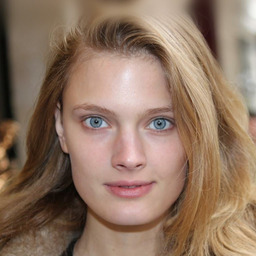} &
\includegraphics[width=0.12\linewidth]{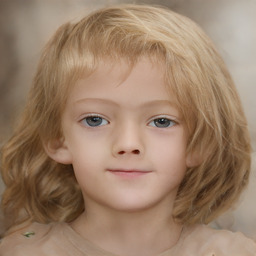} &
\includegraphics[width=0.12\linewidth]{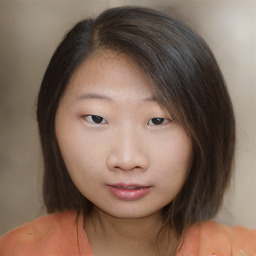} &
\includegraphics[width=0.12\linewidth]{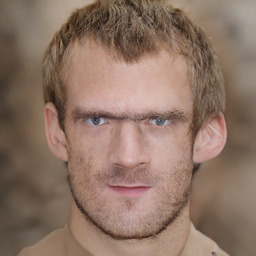} &
\includegraphics[width=0.12\linewidth]{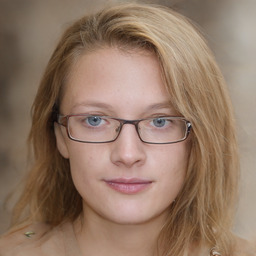} &
\includegraphics[width=0.12\linewidth]{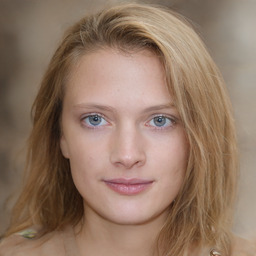} &
\includegraphics[width=0.12\linewidth]{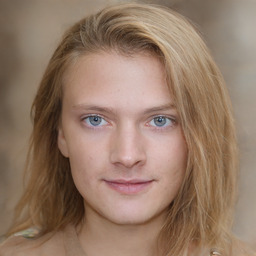} &
\includegraphics[width=0.12\linewidth]{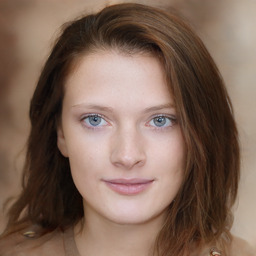} \\

\begin{turn}{90} ~~~ LORD \end{turn} &
\includegraphics[width=0.12\linewidth]{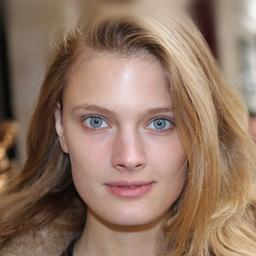} &
\includegraphics[width=0.12\linewidth]{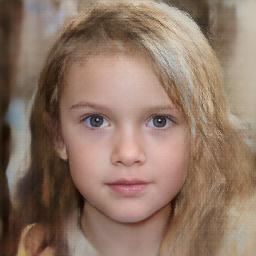} &
\includegraphics[width=0.12\linewidth]{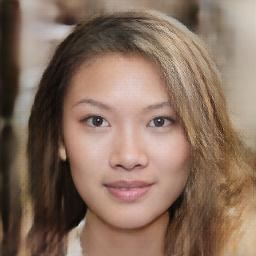} &
\includegraphics[width=0.12\linewidth]{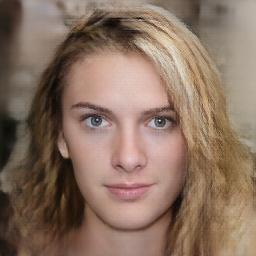} &
\includegraphics[width=0.12\linewidth]{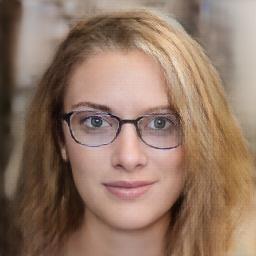} &
\includegraphics[width=0.12\linewidth]{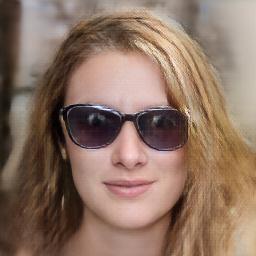} &
\includegraphics[width=0.12\linewidth]{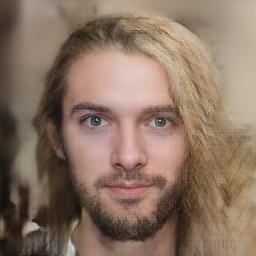} &
\includegraphics[width=0.12\linewidth]{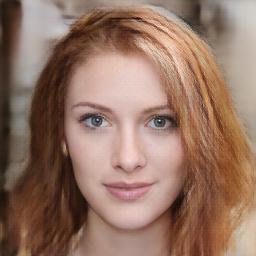} \\

\begin{turn}{90} ~~~~ \textbf{Ours} \end{turn} &
\includegraphics[width=0.12\linewidth]{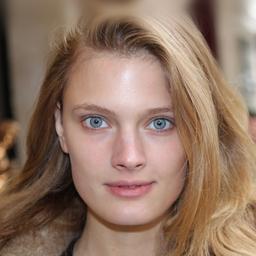} &
\includegraphics[width=0.12\linewidth]{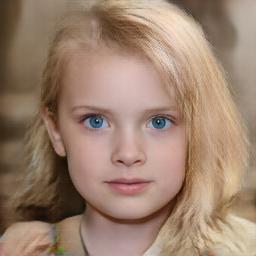} &
\includegraphics[width=0.12\linewidth]{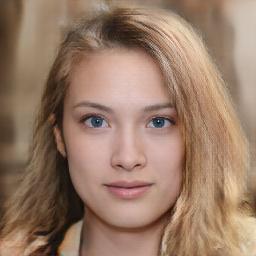} &
\includegraphics[width=0.12\linewidth]{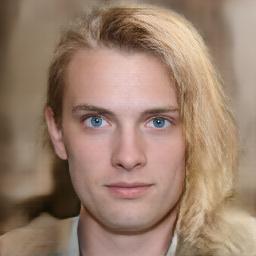} &
\includegraphics[width=0.12\linewidth]{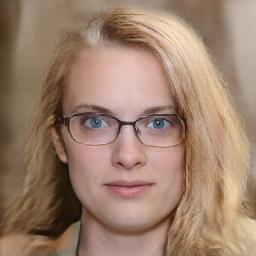} &
\includegraphics[width=0.12\linewidth]{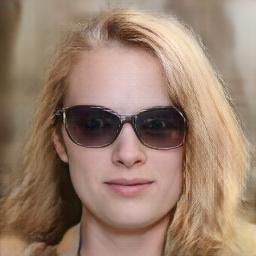} &
\includegraphics[width=0.12\linewidth]{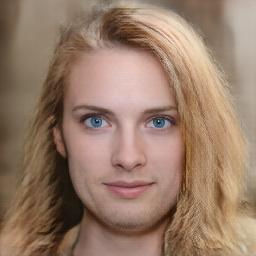} &
\includegraphics[width=0.12\linewidth]{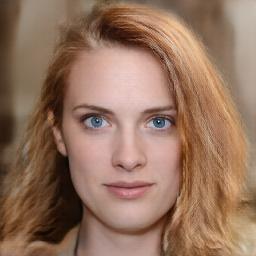} \\

\end{tabular}
\end{center}
\caption{Zero-shot manipulation of human faces. StyleGAN-based approaches  (TediGAN and StyleCLIP) mainly disentangle highly-localized visual concepts (e.g. beard) while global concepts (e.g. gender) seem to be entangled with identity and expression. Moreover, their manipulation requires manual calibration, leading to negligible changes (e.g. invisible glasses) or extreme edits (e.g. translation to asian does not preserve identity). LORD does not require calibration but struggles to disentangle attributes which are not perfectly uncorrelated (e.g. the gender attribute is ignored and remains entangled with beard and hair color). Our method generates highly disentangled results without manual tuning. Note that all manipulations are subject to attribute correlation e.g. beard is not significantly added to females.}
\label{fig:ffhq_additional2}
\end{figure*}

\begin{figure*}[t]
\begin{center}
\begin{tabular}{@{\hskip0pt}c@{\hskip2pt}c@{\hskip3pt}c@{\hskip0pt}c@{\hskip0pt}c@{\hskip0pt}c@{\hskip0pt}c@{\hskip0pt}c@{\hskip0pt}c}

& Input & Kid & Asian & Gender & Glasses & Shades & Beard & Red hair \\

\begin{turn}{90} ~ TediGAN \end{turn} &
\includegraphics[width=0.12\linewidth]{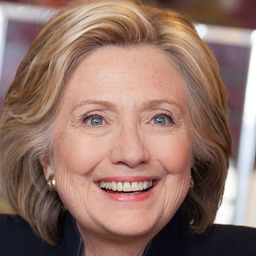} &
\includegraphics[width=0.12\linewidth]{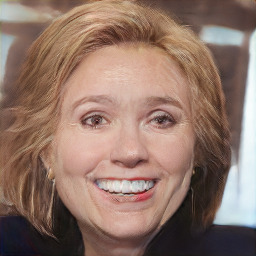} &
\includegraphics[width=0.12\linewidth]{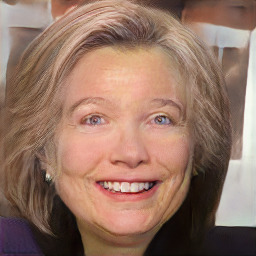} &
\includegraphics[width=0.12\linewidth]{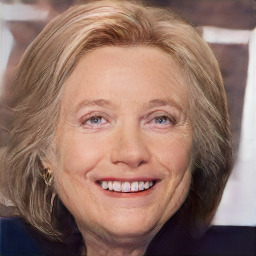} &
\includegraphics[width=0.12\linewidth]{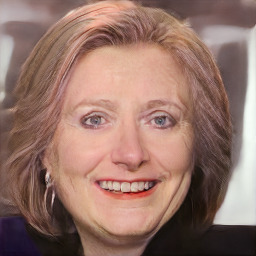} &
\includegraphics[width=0.12\linewidth]{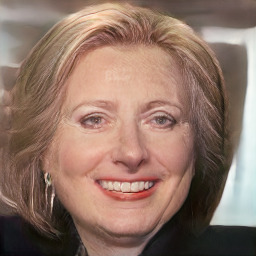} &
\includegraphics[width=0.12\linewidth]{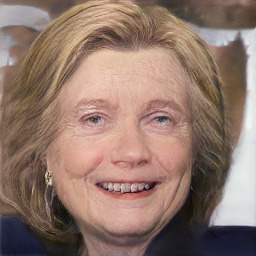} &
\includegraphics[width=0.12\linewidth]{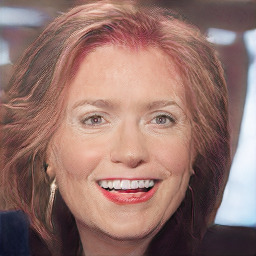} \\

\begin{turn}{90} ~ \footnotesize StyleCLIP- \end{turn} &
\includegraphics[width=0.12\linewidth]{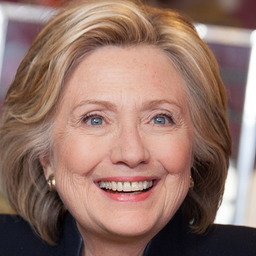} &
\includegraphics[width=0.12\linewidth]{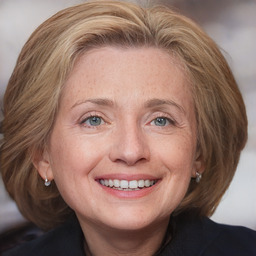} &
\includegraphics[width=0.12\linewidth]{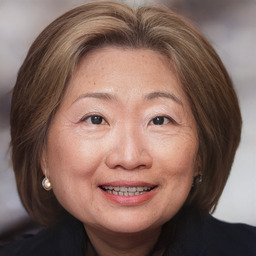} &
\includegraphics[width=0.12\linewidth]{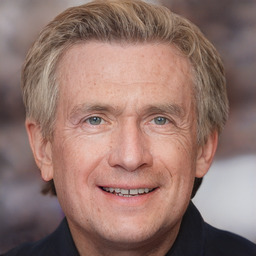} &
\includegraphics[width=0.12\linewidth]{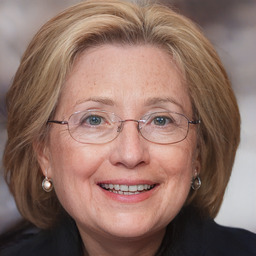} &
\includegraphics[width=0.12\linewidth]{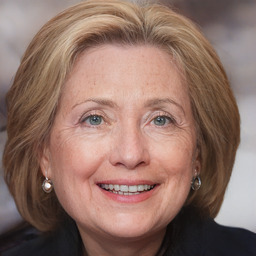} &
\includegraphics[width=0.12\linewidth]{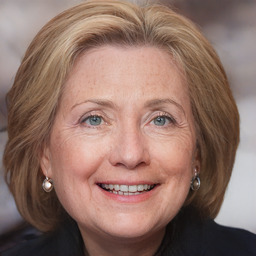} &
\includegraphics[width=0.12\linewidth]{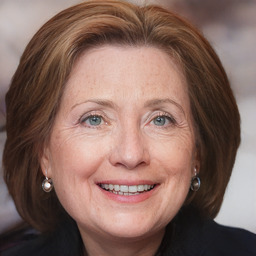} \\

\begin{turn}{90} \footnotesize ~ StyleCLIP+ \end{turn} &
\includegraphics[width=0.12\linewidth]{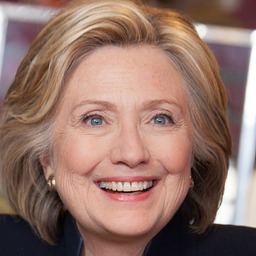} &
\includegraphics[width=0.12\linewidth]{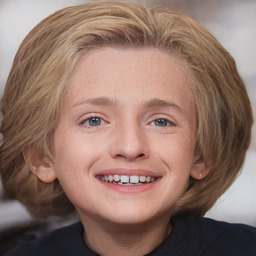} &
\includegraphics[width=0.12\linewidth]{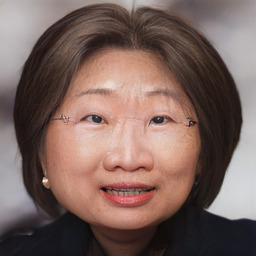} &
\includegraphics[width=0.12\linewidth]{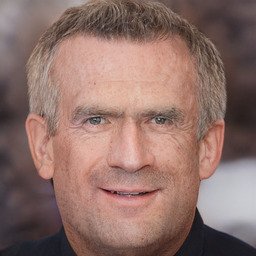} &
\includegraphics[width=0.12\linewidth]{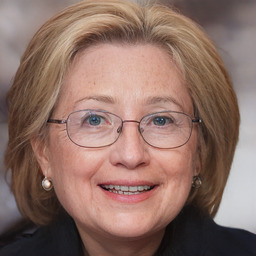} &
\includegraphics[width=0.12\linewidth]{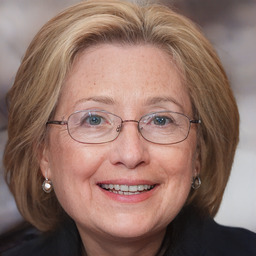} &
\includegraphics[width=0.12\linewidth]{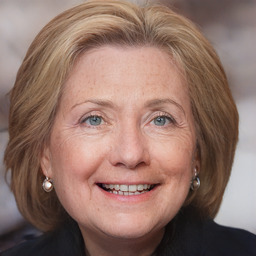} &
\includegraphics[width=0.12\linewidth]{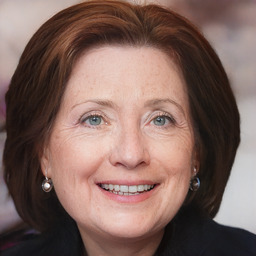} \\

\begin{turn}{90} ~~~ LORD \end{turn} &
\includegraphics[width=0.12\linewidth]{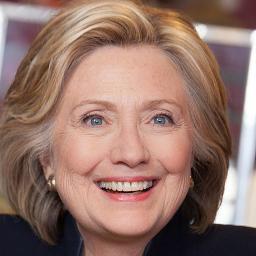} &
\includegraphics[width=0.12\linewidth]{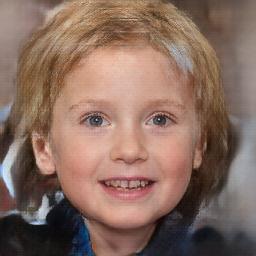} &
\includegraphics[width=0.12\linewidth]{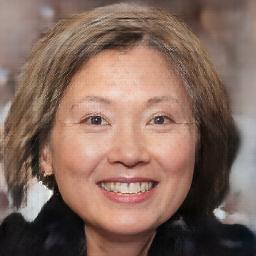} &
\includegraphics[width=0.12\linewidth]{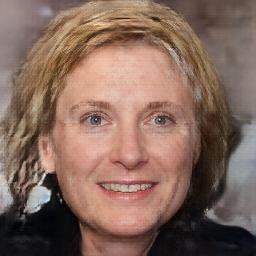} &
\includegraphics[width=0.12\linewidth]{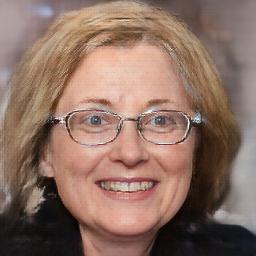} &
\includegraphics[width=0.12\linewidth]{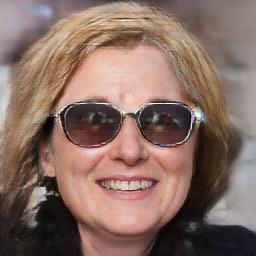} &
\includegraphics[width=0.12\linewidth]{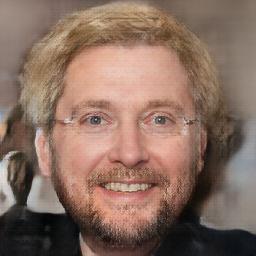} &
\includegraphics[width=0.12\linewidth]{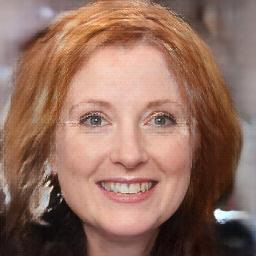} \\

\begin{turn}{90} ~~~~ \textbf{Ours} \end{turn} &
\includegraphics[width=0.12\linewidth]{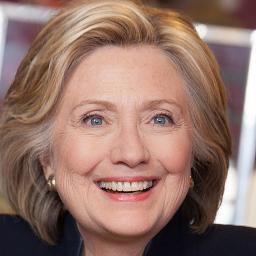} &
\includegraphics[width=0.12\linewidth]{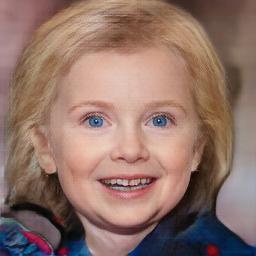} &
\includegraphics[width=0.12\linewidth]{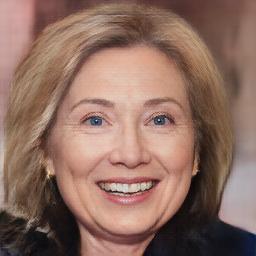} &
\includegraphics[width=0.12\linewidth]{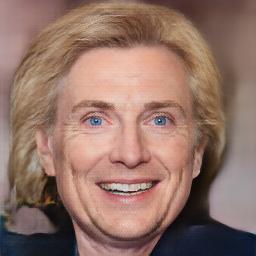} &
\includegraphics[width=0.12\linewidth]{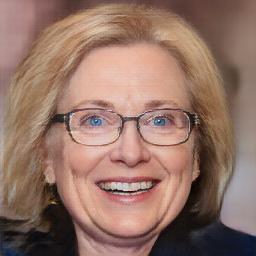} &
\includegraphics[width=0.12\linewidth]{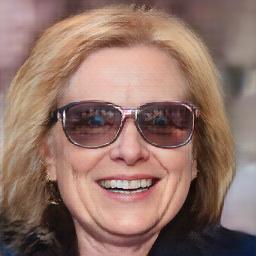} &
\includegraphics[width=0.12\linewidth]{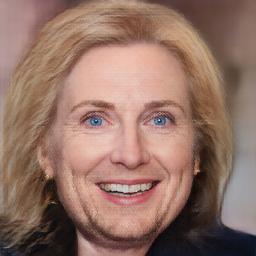} &
\includegraphics[width=0.12\linewidth]{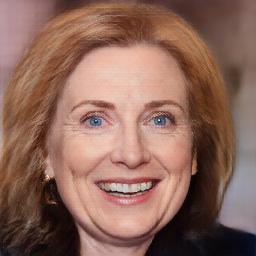} \\

\begin{turn}{90} ~ TediGAN \end{turn} &
\includegraphics[width=0.12\linewidth]{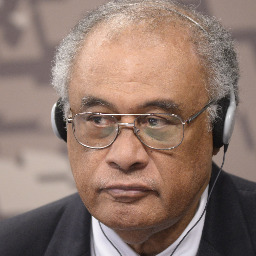} &
\includegraphics[width=0.12\linewidth]{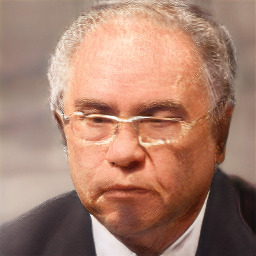} &
\includegraphics[width=0.12\linewidth]{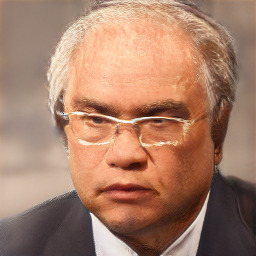} &
\includegraphics[width=0.12\linewidth]{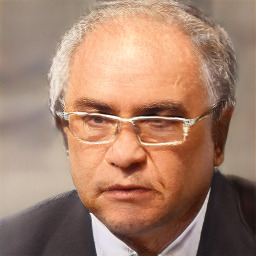} &
\includegraphics[width=0.12\linewidth]{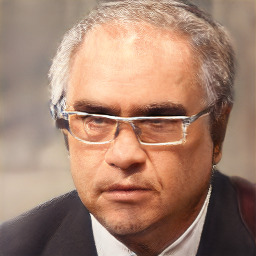} &
\includegraphics[width=0.12\linewidth]{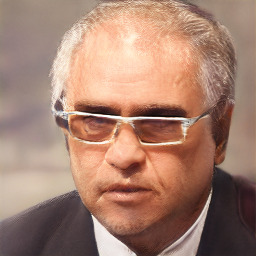} &
\includegraphics[width=0.12\linewidth]{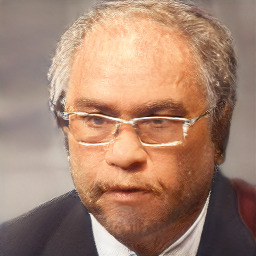} &
\includegraphics[width=0.12\linewidth]{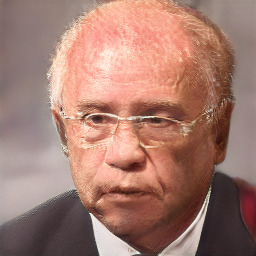} \\

\begin{turn}{90} ~ \footnotesize StyleCLIP- \end{turn} &
\includegraphics[width=0.12\linewidth]{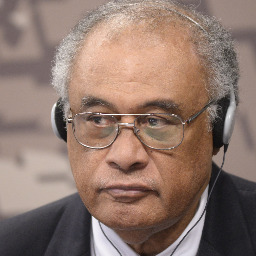} &
\includegraphics[width=0.12\linewidth]{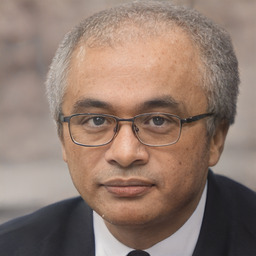} &
\includegraphics[width=0.12\linewidth]{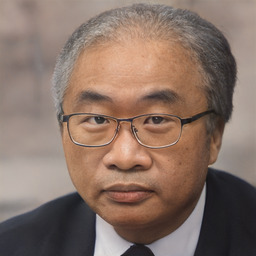} &
\includegraphics[width=0.12\linewidth]{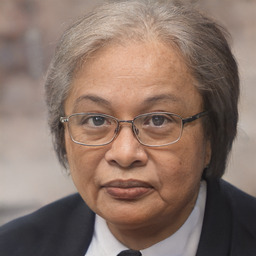} &
\includegraphics[width=0.12\linewidth]{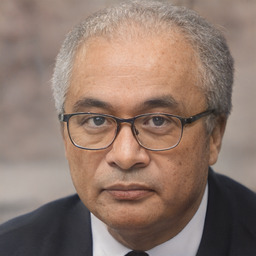} &
\includegraphics[width=0.12\linewidth]{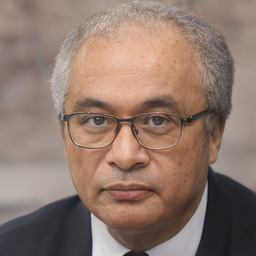} &
\includegraphics[width=0.12\linewidth]{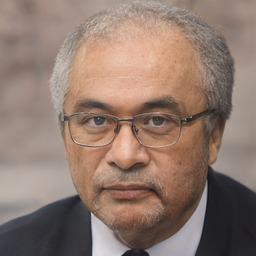} &
\includegraphics[width=0.12\linewidth]{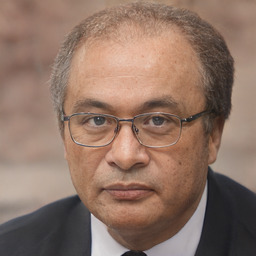} \\

\begin{turn}{90} \footnotesize ~ StyleCLIP+ \end{turn} &
\includegraphics[width=0.12\linewidth]{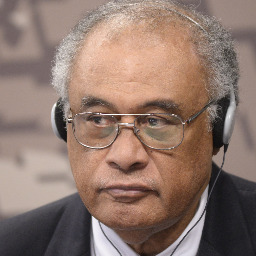} &
\includegraphics[width=0.12\linewidth]{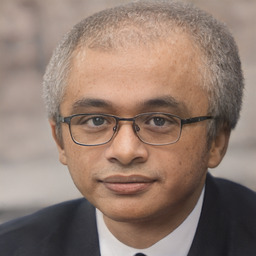} &
\includegraphics[width=0.12\linewidth]{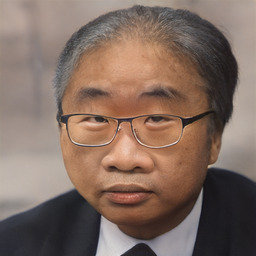} &
\includegraphics[width=0.12\linewidth]{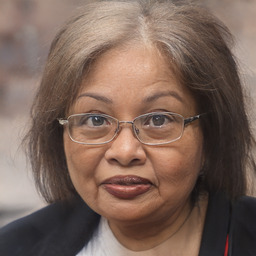} &
\includegraphics[width=0.12\linewidth]{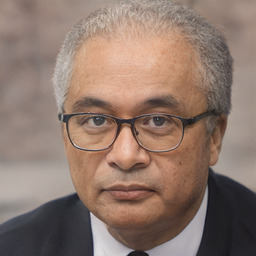} &
\includegraphics[width=0.12\linewidth]{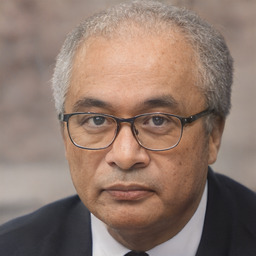} &
\includegraphics[width=0.12\linewidth]{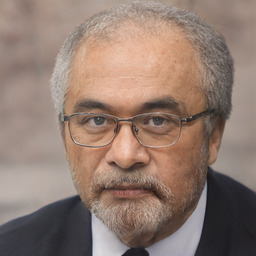} &
\includegraphics[width=0.12\linewidth]{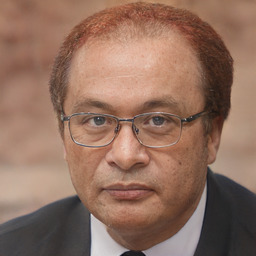} \\

\begin{turn}{90} ~~~ LORD \end{turn} &
\includegraphics[width=0.12\linewidth]{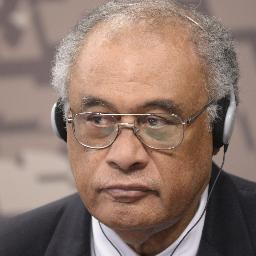} &
\includegraphics[width=0.12\linewidth]{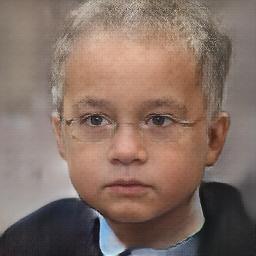} &
\includegraphics[width=0.12\linewidth]{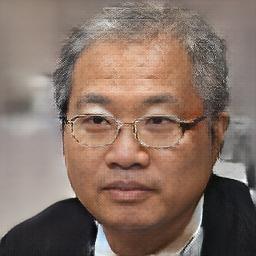} &
\includegraphics[width=0.12\linewidth]{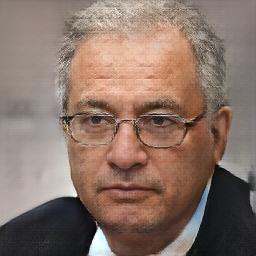} &
\includegraphics[width=0.12\linewidth]{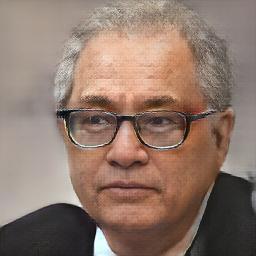} &
\includegraphics[width=0.12\linewidth]{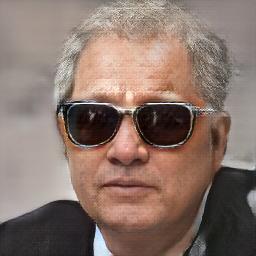} &
\includegraphics[width=0.12\linewidth]{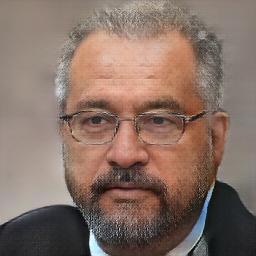} &
\includegraphics[width=0.12\linewidth]{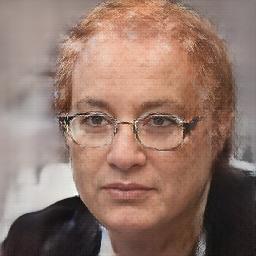} \\

\begin{turn}{90} ~~~~ \textbf{Ours} \end{turn} &
\includegraphics[width=0.12\linewidth]{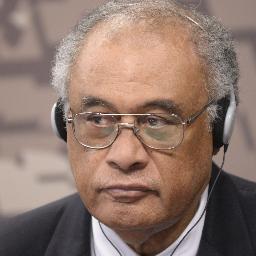} &
\includegraphics[width=0.12\linewidth]{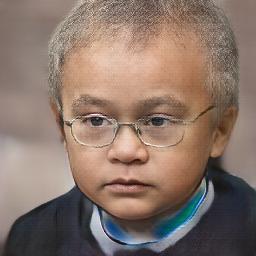} &
\includegraphics[width=0.12\linewidth]{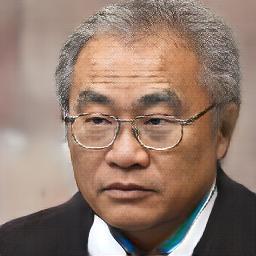} &
\includegraphics[width=0.12\linewidth]{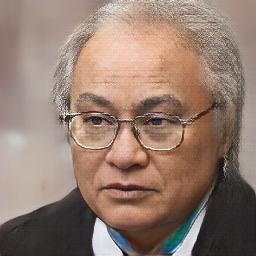} &
\includegraphics[width=0.12\linewidth]{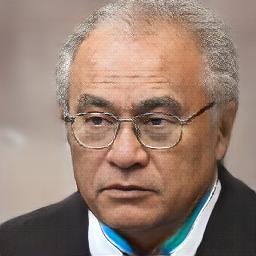} &
\includegraphics[width=0.12\linewidth]{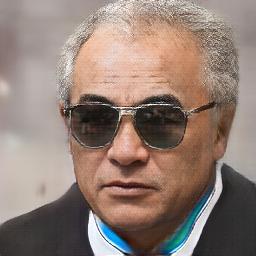} &
\includegraphics[width=0.12\linewidth]{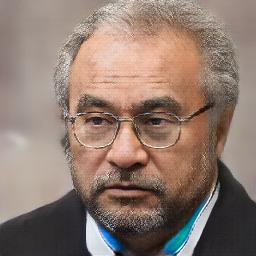} &
\includegraphics[width=0.12\linewidth]{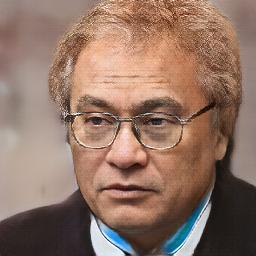} \\

\end{tabular}
\end{center}
\caption{Zero-shot manipulation of human faces. StyleGAN-based approaches  (TediGAN and StyleCLIP) mainly disentangle highly-localized visual concepts (e.g. beard) while global concepts (e.g. gender) seem to be entangled with identity and expression. Moreover, their manipulation requires manual calibration, leading to negligible changes (e.g. invisible glasses) or extreme edits (e.g. translation to asian does not preserve identity). LORD does not require calibration but struggles to disentangle attributes which are not perfectly uncorrelated (e.g. the gender attribute is ignored and remains entangled with beard and hair color). Our method generates highly disentangled results without manual tuning. Note that all manipulations are subject to attribute correlation e.g. beard is not significantly added to females.}
\label{fig:ffhq_additional3}
\end{figure*}

Tab.~\ref{tab:real_images_quantitative_ad} shows the AD scores and manipulation strength of all methods while manipulating different attributes of interest. We stress that quantitative measurements of this sort are not perfect and can sometimes be misleading. However, let us briefly review the main trends reflected by these metrics (the same trends are clearly visualized in Fig.~\ref{fig:ffhq_additional1},\ref{fig:ffhq_additional2},\ref{fig:ffhq_additional3}): (i) StyleCLIP tends to over manipulate the desired attribute and causes changes to other attributes of the input image, resulting in inferior disentanglement and leading to higher AD scores. This can be clearly seen when changing gender. (ii) LORD struggles to disentangle attributes which are not perfectly uncorrelated e.g. manipulating gender does not affect the input image at all (low manipulation strength which results in a misleading low AD score) while adding beard to females leads to gender swapping (higher AD scores). Note that TediGAN mostly introduces artifacts without manipulating the desired attribute (low manipulation strengths), and therefore maintains misleading low AD scores. The manipulation strength of the ethnicity attribute is not reported due to the lack of a pretrained classifier, and the manipulation strength of beard is not assessed as this attribute is correlated with other attributes (e.g. gender) and the manipulation should not cause any effect in many cases.

\begin{table}
  \caption{Evaluation of disentanglement measured by Attribute Dependency ($\downarrow$) and [manipulation strength ($\uparrow$)], on real human face images.}
  \label{tab:real_images_quantitative_ad}
  \centering
  \begin{tabular}{lcccccc}
    \toprule
     & Age & Beard & Ethnicity & Gender & Glasses & Hair Color \\
    \midrule
    TediGAN \cite{xia2021towards} & 0.39 [0.04] & 0.38 [-] & 0.41 [-] & 0.40 [0.02] & 0.31 [0.18] & 0.37 [0.28] \\
    StyleCLIP \cite{patashnik2021styleclip} & 0.45 [-0.07] & 0.42 [-] & 0.40 [-] & 0.78 [0.57] & 0.35 [0.22] & 0.44 [0.17] \\
    LORD \cite{gabbay2020lord} & 0.41 [0.13] & 0.65 [-] & 0.38 [-] & 0.36 [0.04] & 0.46 [0.19] & 0.38 [0.26] \\
    Ours & 0.40 [0.12] & 0.36 [-] & 0.40 [-] & 0.44 [0.20] & 0.49 [0.23] & 0.37 [0.28] \\
    \bottomrule
  \end{tabular}
\end{table}

\begin{figure*}[t]
\begin{center}
\begin{tabular}{c@{\hskip1pt}c@{\hskip0pt}c@{\hskip0pt}c@{\hskip0pt}c@{\hskip0pt}c@{\hskip0pt}c@{\hskip0pt}c}

\footnotesize Input & \footnotesize Boerboel & \footnotesize Labradoodle & \footnotesize Husky & \footnotesize Chihuahua & \footnotesize Cheetah & \footnotesize Jaguar & \footnotesize Bombay cat \\

\includegraphics[width=0.115\linewidth]{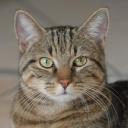} &
\includegraphics[width=0.115\linewidth]{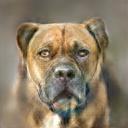} &
\includegraphics[width=0.115\linewidth]{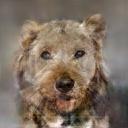} &
\includegraphics[width=0.115\linewidth]{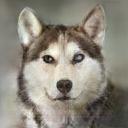} &
\includegraphics[width=0.115\linewidth]{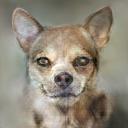} &
\includegraphics[width=0.115\linewidth]{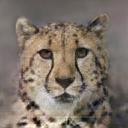} &
\includegraphics[width=0.115\linewidth]{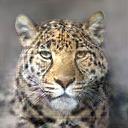} &
\includegraphics[width=0.115\linewidth]{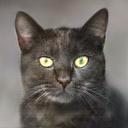} \\

\includegraphics[width=0.115\linewidth]{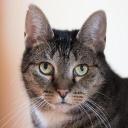} &
\includegraphics[width=0.115\linewidth]{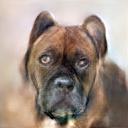} &
\includegraphics[width=0.115\linewidth]{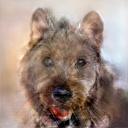} &
\includegraphics[width=0.115\linewidth]{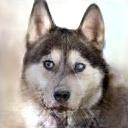} &
\includegraphics[width=0.115\linewidth]{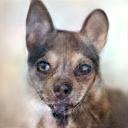} &
\includegraphics[width=0.115\linewidth]{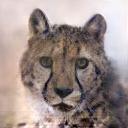} &
\includegraphics[width=0.115\linewidth]{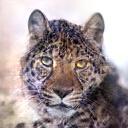} &
\includegraphics[width=0.115\linewidth]{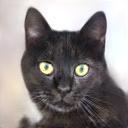} \\

\includegraphics[width=0.115\linewidth]{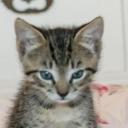} &
\includegraphics[width=0.115\linewidth]{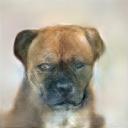} &
\includegraphics[width=0.115\linewidth]{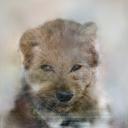} &
\includegraphics[width=0.115\linewidth]{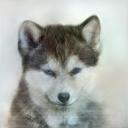} &
\includegraphics[width=0.115\linewidth]{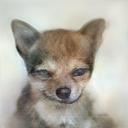} &
\includegraphics[width=0.115\linewidth]{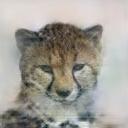} &
\includegraphics[width=0.115\linewidth]{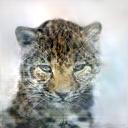} &
\includegraphics[width=0.115\linewidth]{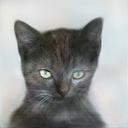} \\

\includegraphics[width=0.115\linewidth]{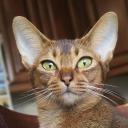} &
\includegraphics[width=0.115\linewidth]{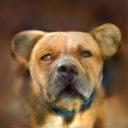} &
\includegraphics[width=0.115\linewidth]{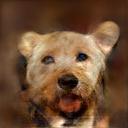} &
\includegraphics[width=0.115\linewidth]{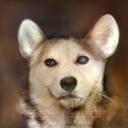} &
\includegraphics[width=0.115\linewidth]{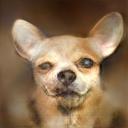} &
\includegraphics[width=0.115\linewidth]{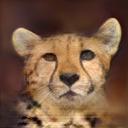} &
\includegraphics[width=0.115\linewidth]{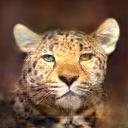} &
\includegraphics[width=0.115\linewidth]{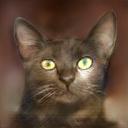} \\

\includegraphics[width=0.115\linewidth]{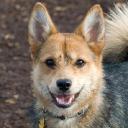} &
\includegraphics[width=0.115\linewidth]{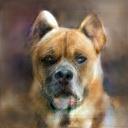} &
\includegraphics[width=0.115\linewidth]{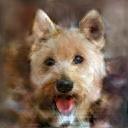} &
\includegraphics[width=0.115\linewidth]{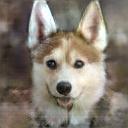} &
\includegraphics[width=0.115\linewidth]{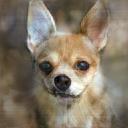} &
\includegraphics[width=0.115\linewidth]{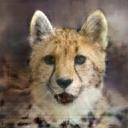} &
\includegraphics[width=0.115\linewidth]{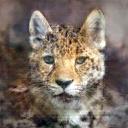} &
\includegraphics[width=0.115\linewidth]{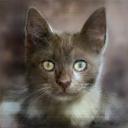} \\

\includegraphics[width=0.115\linewidth]{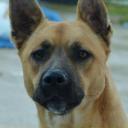} &
\includegraphics[width=0.115\linewidth]{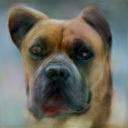} &
\includegraphics[width=0.115\linewidth]{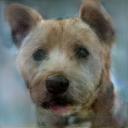} &
\includegraphics[width=0.115\linewidth]{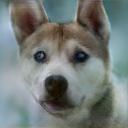} &
\includegraphics[width=0.115\linewidth]{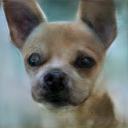} &
\includegraphics[width=0.115\linewidth]{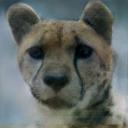} &
\includegraphics[width=0.115\linewidth]{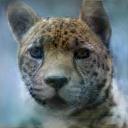} &
\includegraphics[width=0.115\linewidth]{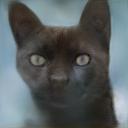} \\

\includegraphics[width=0.115\linewidth]{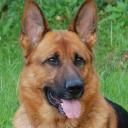} &
\includegraphics[width=0.115\linewidth]{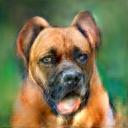} &
\includegraphics[width=0.115\linewidth]{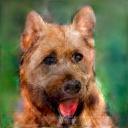} &
\includegraphics[width=0.115\linewidth]{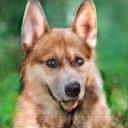} &
\includegraphics[width=0.115\linewidth]{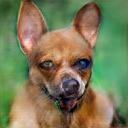} &
\includegraphics[width=0.115\linewidth]{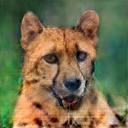} &
\includegraphics[width=0.115\linewidth]{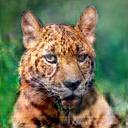} &
\includegraphics[width=0.115\linewidth]{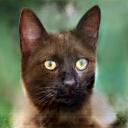} \\

\includegraphics[width=0.115\linewidth]{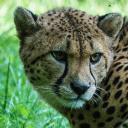} &
\includegraphics[width=0.115\linewidth]{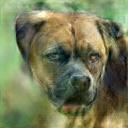} &
\includegraphics[width=0.115\linewidth]{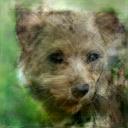} &
\includegraphics[width=0.115\linewidth]{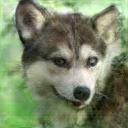} &
\includegraphics[width=0.115\linewidth]{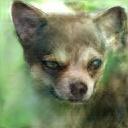} &
\includegraphics[width=0.115\linewidth]{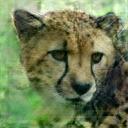} &
\includegraphics[width=0.115\linewidth]{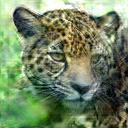} &
\includegraphics[width=0.115\linewidth]{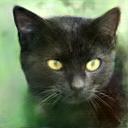} \\

\includegraphics[width=0.115\linewidth]{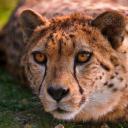} &
\includegraphics[width=0.115\linewidth]{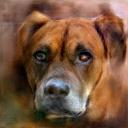} &
\includegraphics[width=0.115\linewidth]{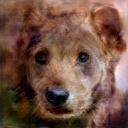} &
\includegraphics[width=0.115\linewidth]{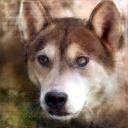} &
\includegraphics[width=0.115\linewidth]{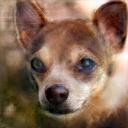} &
\includegraphics[width=0.115\linewidth]{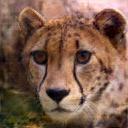} &
\includegraphics[width=0.115\linewidth]{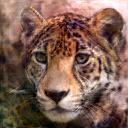} &
\includegraphics[width=0.115\linewidth]{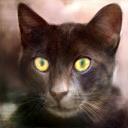} \\

\includegraphics[width=0.115\linewidth]{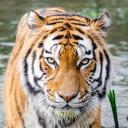} &
\includegraphics[width=0.115\linewidth]{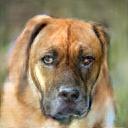} &
\includegraphics[width=0.115\linewidth]{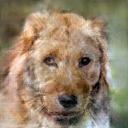} &
\includegraphics[width=0.115\linewidth]{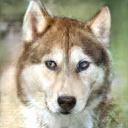} &
\includegraphics[width=0.115\linewidth]{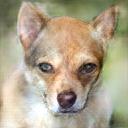} &
\includegraphics[width=0.115\linewidth]{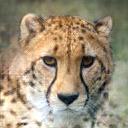} &
\includegraphics[width=0.115\linewidth]{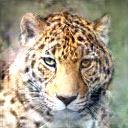} &
\includegraphics[width=0.115\linewidth]{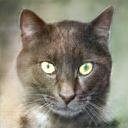} \\

\includegraphics[width=0.115\linewidth]{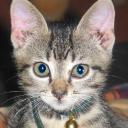} &
\includegraphics[width=0.115\linewidth]{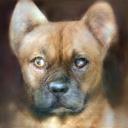} &
\includegraphics[width=0.115\linewidth]{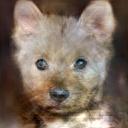} &
\includegraphics[width=0.115\linewidth]{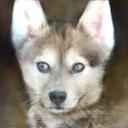} &
\includegraphics[width=0.115\linewidth]{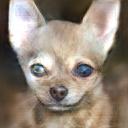} &
\includegraphics[width=0.115\linewidth]{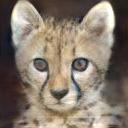} &
\includegraphics[width=0.115\linewidth]{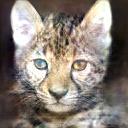} &
\includegraphics[width=0.115\linewidth]{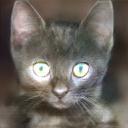} \\



\end{tabular}
\end{center}
\caption{More zero-shot translation results of animal species on AFHQ.}
\label{fig:afhq_additional}
\end{figure*}

\begin{figure*}[t]
\begin{center}
\begin{tabular}{c@{\hskip1pt}c@{\hskip0pt}c@{\hskip0pt}c@{\hskip0pt}c@{\hskip0pt}c@{\hskip0pt}c@{\hskip0pt}c@{\hskip0pt}c}

Input & Jeep & Sports & Family & Black & White & Blue & Red & Yellow \\

\includegraphics[width=0.105\linewidth]{figures/cars/j/input.jpg} &
\includegraphics[width=0.105\linewidth]{figures/cars/j/jeep.jpg} &
\includegraphics[width=0.105\linewidth]{figures/cars/j/sports.jpg} &
\includegraphics[width=0.105\linewidth]{figures/cars/j/family.jpg} &
\includegraphics[width=0.105\linewidth]{figures/cars/j/black.jpg} &
\includegraphics[width=0.105\linewidth]{figures/cars/j/white.jpg} &
\includegraphics[width=0.105\linewidth]{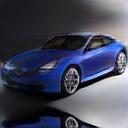} &
\includegraphics[width=0.105\linewidth]{figures/cars/j/red.jpg} &
\includegraphics[width=0.105\linewidth]{figures/cars/j/yellow.jpg} \\

\includegraphics[width=0.105\linewidth]{figures/cars/b/input.jpg} &
\includegraphics[width=0.105\linewidth]{figures/cars/b/jeep.jpg} &
\includegraphics[width=0.105\linewidth]{figures/cars/b/sports.jpg} &
\includegraphics[width=0.105\linewidth]{figures/cars/b/family.jpg} &
\includegraphics[width=0.105\linewidth]{figures/cars/b/black.jpg} &
\includegraphics[width=0.105\linewidth]{figures/cars/b/white.jpg} &
\includegraphics[width=0.105\linewidth]{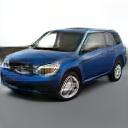} &
\includegraphics[width=0.105\linewidth]{figures/cars/b/red.jpg} &
\includegraphics[width=0.105\linewidth]{figures/cars/b/yellow.jpg} \\

\includegraphics[width=0.105\linewidth]{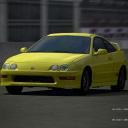} &
\includegraphics[width=0.105\linewidth]{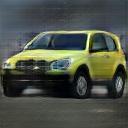} &
\includegraphics[width=0.105\linewidth]{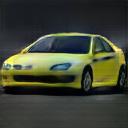} &
\includegraphics[width=0.105\linewidth]{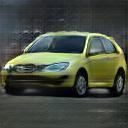} &
\includegraphics[width=0.105\linewidth]{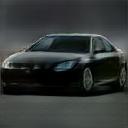} &
\includegraphics[width=0.105\linewidth]{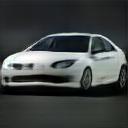} &
\includegraphics[width=0.105\linewidth]{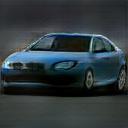} &
\includegraphics[width=0.105\linewidth]{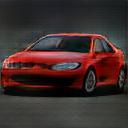} &
\includegraphics[width=0.105\linewidth]{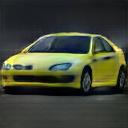} \\

\includegraphics[width=0.105\linewidth]{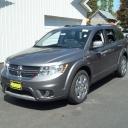} &
\includegraphics[width=0.105\linewidth]{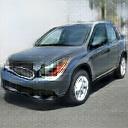} &
\includegraphics[width=0.105\linewidth]{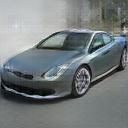} &
\includegraphics[width=0.105\linewidth]{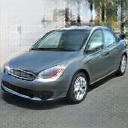} &
\includegraphics[width=0.105\linewidth]{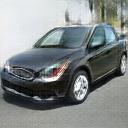} &
\includegraphics[width=0.105\linewidth]{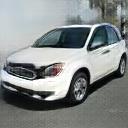} &
\includegraphics[width=0.105\linewidth]{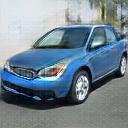} &
\includegraphics[width=0.105\linewidth]{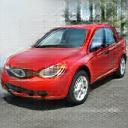} &
\includegraphics[width=0.105\linewidth]{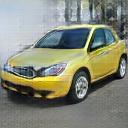} \\

\includegraphics[width=0.105\linewidth]{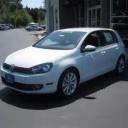} &
\includegraphics[width=0.105\linewidth]{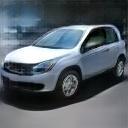} &
\includegraphics[width=0.105\linewidth]{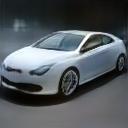} &
\includegraphics[width=0.105\linewidth]{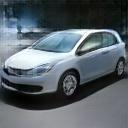} &
\includegraphics[width=0.105\linewidth]{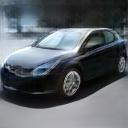} &
\includegraphics[width=0.105\linewidth]{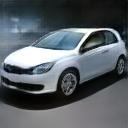} &
\includegraphics[width=0.105\linewidth]{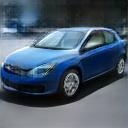} &
\includegraphics[width=0.105\linewidth]{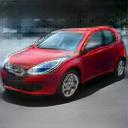} &
\includegraphics[width=0.105\linewidth]{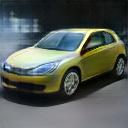} \\

\includegraphics[width=0.105\linewidth]{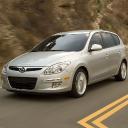} &
\includegraphics[width=0.105\linewidth]{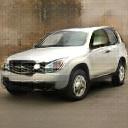} &
\includegraphics[width=0.105\linewidth]{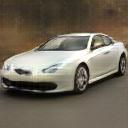} &
\includegraphics[width=0.105\linewidth]{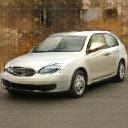} &
\includegraphics[width=0.105\linewidth]{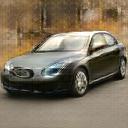} &
\includegraphics[width=0.105\linewidth]{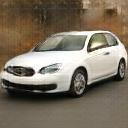} &
\includegraphics[width=0.105\linewidth]{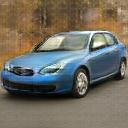} &
\includegraphics[width=0.105\linewidth]{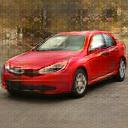} &
\includegraphics[width=0.105\linewidth]{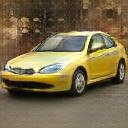} \\

\includegraphics[width=0.105\linewidth]{figures/cars/i/input.jpg} &
\includegraphics[width=0.105\linewidth]{figures/cars/i/jeep.jpg} &
\includegraphics[width=0.105\linewidth]{figures/cars/i/sports.jpg} &
\includegraphics[width=0.105\linewidth]{figures/cars/i/family.jpg} &
\includegraphics[width=0.105\linewidth]{figures/cars/i/black.jpg} &
\includegraphics[width=0.105\linewidth]{figures/cars/i/white.jpg} &
\includegraphics[width=0.105\linewidth]{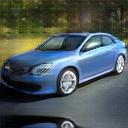} &
\includegraphics[width=0.105\linewidth]{figures/cars/i/red.jpg} &
\includegraphics[width=0.105\linewidth]{figures/cars/i/yellow.jpg} \\

\includegraphics[width=0.105\linewidth]{figures/cars/k/input.jpg} &
\includegraphics[width=0.105\linewidth]{figures/cars/k/jeep.jpg} &
\includegraphics[width=0.105\linewidth]{figures/cars/k/sports.jpg} &
\includegraphics[width=0.105\linewidth]{figures/cars/k/family.jpg} &
\includegraphics[width=0.105\linewidth]{figures/cars/k/black.jpg} &
\includegraphics[width=0.105\linewidth]{figures/cars/k/white.jpg} &
\includegraphics[width=0.105\linewidth]{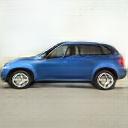} &
\includegraphics[width=0.105\linewidth]{figures/cars/k/red.jpg} &
\includegraphics[width=0.105\linewidth]{figures/cars/k/yellow.jpg} \\

\includegraphics[width=0.105\linewidth]{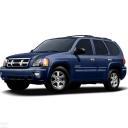} &
\includegraphics[width=0.105\linewidth]{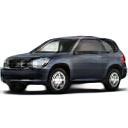} &
\includegraphics[width=0.105\linewidth]{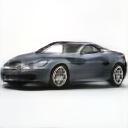} &
\includegraphics[width=0.105\linewidth]{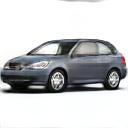} &
\includegraphics[width=0.105\linewidth]{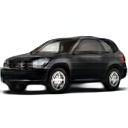} &
\includegraphics[width=0.105\linewidth]{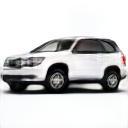} &
\includegraphics[width=0.105\linewidth]{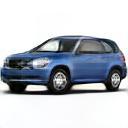} &
\includegraphics[width=0.105\linewidth]{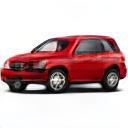} &
\includegraphics[width=0.105\linewidth]{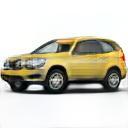} \\

\includegraphics[width=0.105\linewidth]{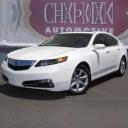} &
\includegraphics[width=0.105\linewidth]{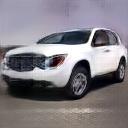} &
\includegraphics[width=0.105\linewidth]{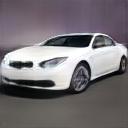} &
\includegraphics[width=0.105\linewidth]{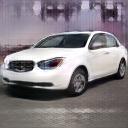} &
\includegraphics[width=0.105\linewidth]{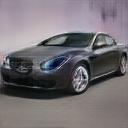} &
\includegraphics[width=0.105\linewidth]{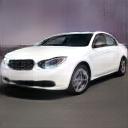} &
\includegraphics[width=0.105\linewidth]{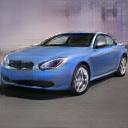} &
\includegraphics[width=0.105\linewidth]{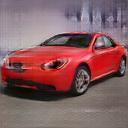} &
\includegraphics[width=0.105\linewidth]{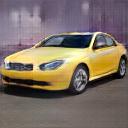} \\

\includegraphics[width=0.105\linewidth]{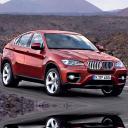} &
\includegraphics[width=0.105\linewidth]{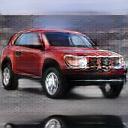} &
\includegraphics[width=0.105\linewidth]{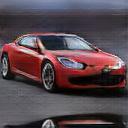} &
\includegraphics[width=0.105\linewidth]{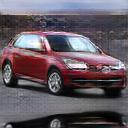} &
\includegraphics[width=0.105\linewidth]{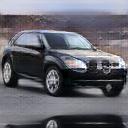} &
\includegraphics[width=0.105\linewidth]{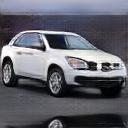} &
\includegraphics[width=0.105\linewidth]{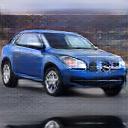} &
\includegraphics[width=0.105\linewidth]{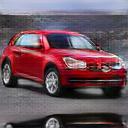} &
\includegraphics[width=0.105\linewidth]{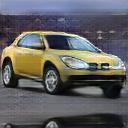} \\

\includegraphics[width=0.105\linewidth]{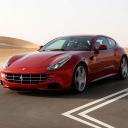} &
\includegraphics[width=0.105\linewidth]{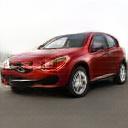} &
\includegraphics[width=0.105\linewidth]{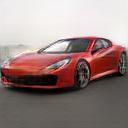} &
\includegraphics[width=0.105\linewidth]{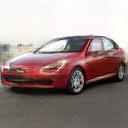} &
\includegraphics[width=0.105\linewidth]{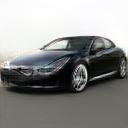} &
\includegraphics[width=0.105\linewidth]{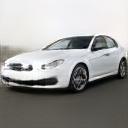} &
\includegraphics[width=0.105\linewidth]{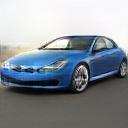} &
\includegraphics[width=0.105\linewidth]{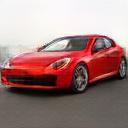} &
\includegraphics[width=0.105\linewidth]{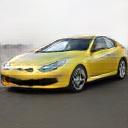} \\

\end{tabular}
\end{center}
\caption{Zero-shot translation results of car types and colors.}
\label{fig:cars_additional}
\end{figure*}

\paragraph{Qualitative Visualizations}
We provide more qualitative results on FFHQ (Fig.~\ref{fig:ffhq_additional1},\ref{fig:ffhq_additional2},\ref{fig:ffhq_additional3}) along with a comparison to TediGAN, StyleCLIP and LORD. More qualitative results on AFHQ and Cars are shown in Fig.~\ref{fig:afhq_additional} and Fig.~\ref{fig:cars_additional}.

\begin{table}[t]
  \caption{Accuracy of factor predictions from the residual code on the synthetic  benchmarks, using 1000 [or 100] labels per attribute of interest. We indicate beside each attribute its number of values. Lower accuracy in predicting the \textcolor{red}{attributes of interest} and higher accuracy in predicting the \textcolor{forestgreen}{residual attributes} indicate better disentanglement.}
  \label{tab:synthetic_quantitative_residual}
  \centering
  \begin{tabular}{lccccc}
    \toprule
    \multicolumn{6}{l}{\textbf{Dataset:} Shapes3D~~\textbf{Attributes:} floor, wall, object~~\textbf{Residuals:} scale, shape, azimuth} \\
    \midrule
	&	\textcolor{red}{floor color [10]		}		&	\textcolor{red}{wall color [10]		}		&	\textcolor{red}{object color [10]		}		\\	\midrule
\citet{locatello2020fewlabels}	&	1.00	[1.00]	&	1.00	[1.00]	&	0.87	[1.00]	\\	
LORD \cite{gabbay2020lord}	&	0.96	[0.80]	&	0.87	[0.75]	&	0.35	[0.49]	\\	
Ours	&	\textbf{0.11}	[\textbf{0.13}]	&	\textbf{0.12}	[\textbf{0.15}]	&	\textbf{0.14}	[\textbf{0.15}]	\\	
\midrule	&	\textcolor{forestgreen}{scale [8]		}		&	\textcolor{forestgreen}{shape [4]		}		&	\textcolor{forestgreen}{azimuth [15]		}		\\	\midrule
\citet{locatello2020fewlabels}	&	0.15	[0.34]	&	0.29	[0.32]	&	0.59	[0.77]	\\	
LORD \cite{gabbay2020lord}	&	0.25	[0.17]	&	0.38	[0.37]	&	0.20	[0.15]	\\	
Ours	&	\textbf{0.75}	[\textbf{0.47}]	&	\textbf{0.97}	[\textbf{0.79}]	&	\textbf{0.79}	[\textbf{0.48}]	\\														
    \midrule
    \multicolumn{6}{l}{\textbf{Dataset:} Cars3D~~\textbf{Attributes:} elevation, azimuth~~\textbf{Residuals:} object} \\
    \midrule
	&	\textcolor{red}{elevation [4]		}		&	\textcolor{red}{azimuth [24]		}		\\	\midrule
\citet{locatello2020fewlabels}	&	0.44	[0.46]	&	0.86	[0.88]	\\	
LORD \cite{gabbay2020lord}	&	0.35	[0.36]	&	0.43	[0.55]	\\	
Ours	&	\textbf{0.29}	[\textbf{0.33}]	&	\textbf{0.28}	[\textbf{0.26}]	\\	
\midrule	&	\textcolor{forestgreen}{object [183]		}		\\	\midrule					
\citet{locatello2020fewlabels}	&	\textbf{0.64}	[\textbf{0.44}]	\\						
LORD \cite{gabbay2020lord}	&	0.37	[0.27]	\\						
Ours	&	0.51	[0.23]	\\						
    \midrule
    \multicolumn{6}{l}{\textbf{Dataset:} dSprites~~\textbf{Attributes:} scale, x, y~~\textbf{Residuals:} orientation, shape} \\
    \midrule
	&	\textcolor{red}{scale [6]		}		&	\textcolor{red}{x [32]		}		&	\textcolor{red}{y [32]		}		\\	\midrule
\citet{locatello2020fewlabels}	&	0.38	[0.35]	&	0.24	[0.20]	&	0.18	[0.34]	\\	
LORD \cite{gabbay2020lord}	&	0.33	[0.31]	&	0.20	[0.15]	&	0.26	[0.18]	\\	
Ours	&	\textbf{0.20}	[\textbf{0.20}]	&	\textbf{0.04}	[\textbf{0.05}]	&	\textbf{0.04}	[\textbf{0.04}]	\\	
\midrule	&	\textcolor{forestgreen}{orientation [40]		}		&	\textcolor{forestgreen}{shape [3]		}		\\	\midrule					
\citet{locatello2020fewlabels}	&	0.04	[0.03]	&	\textbf{0.44}	[\textbf{0.44}]	\\						
LORD \cite{gabbay2020lord}	&	0.03	[0.03]	&	0.43	[0.42]	\\						
Ours	&	\textbf{0.06}	[\textbf{0.06}]	&	0.40	[0.41]	\\										
    \midrule
    \multicolumn{6}{c}{\textbf{Dataset:} SmallNORB~~\textbf{Attributes:} elevation, azimuth, lighting~~\textbf{Residuals:} category, instance} \\
    \midrule
	&	\textcolor{red}{elevation [9]		}		&	\textcolor{red}{azimuth [18]		}		&	\textcolor{red}{lighting [6]		}		\\	\midrule	
\citet{locatello2020fewlabels}	&	\textbf{0.16}	[\textbf{0.17}]	&	0.12	[\textbf{0.10}]	&	0.91	[0.91]	\\		
LORD \cite{gabbay2020lord}	&	\textbf{0.16}	[\textbf{0.17}]	&	0.11	[0.12]	&	0.89	[0.87]	\\		
Ours	&	\textbf{0.16}	[0.18]	&	\textbf{0.10}	[\textbf{0.10}]	&	\textbf{0.24}	[\textbf{0.24}]	\\		
\midrule	&	\textcolor{forestgreen}{category [5]		}		&	\textcolor{forestgreen}{instance [10]		}		\\	\midrule						
\citet{locatello2020fewlabels}	&	0.48	[\textbf{0.54}]	&	0.14	[\textbf{0.14}]	\\							
LORD \cite{gabbay2020lord}	&	0.47	[0.50]	&	0.14	[\textbf{0.14}]	\\							
Ours	&	\textbf{0.59}	[0.41]	&	\textbf{0.16}	[\textbf{0.14}]	\\											
    
    \bottomrule
  \end{tabular}
  \vspace{-0.5em}
\end{table}

\begin{table}
  \caption{Attribute classification accuracy using 1000 [or 100] labels per attribute.}
  \label{tab:ablation_full}
  \centering
  \begin{tabular}{lllll}
    \toprule
    
        Dataset & Attribute [No. of Values]  &	Pseudo-labels				&	Ours				\\
    \midrule
    
\multirow{3}{*}{Shapes3D}	 &					floor color [10]	&	\textbf{1.00}	[0.92]	&	\textbf{1.00}	[\textbf{1.00}]  	\\
& wall color [10]	&	\textbf{1.00}	[0.91]	&	\textbf{1.00}	[\textbf{1.00}] 	\\
& object color [10] 	&	\textbf{1.00}	[0.68]	&	\textbf{1.00}	[\textbf{0.98}] 	\\

\midrule
											
\multirow{2}{*}{Cars3D}	 &											
elevation [4]	&	0.81	[0.51]	&	\textbf{0.85}	[\textbf{0.52}]	\\
& azimuth [24]	&	0.83	[0.40]	&	\textbf{0.85}	[\textbf{0.49}]	\\
											
\midrule

\multirow{3}{*}{dSprites}	 &											
scale [6]	&	0.45	[0.36]	&	\textbf{0.51}	[\textbf{0.47}]	\\
& x [32]	&	0.46	[0.23]	&	\textbf{0.76}	[\textbf{0.33}]	\\
 & y [32]	&	0.46	[0.25]	&	\textbf{0.76}	[\textbf{0.42}]	\\

\midrule

\multirow{3}{*}{SmallNORB}	 &											
elevation [9]	&	\textbf{0.28}	[\textbf{0.19}]	&	\textbf{0.28}	[0.16]	\\
& azimuth [18]	&	0.34	[\textbf{0.11}]	&	\textbf{0.36}	[\textbf{0.11}]	\\
& lighting [6]	&	\textbf{0.92}	[0.86]	&	\textbf{0.92}	[\textbf{0.90}]	\\

    \bottomrule
  \end{tabular}
\end{table}

\subsection{Training Resources}
Training our models on the synthetic datasets takes approximately $3-5$ hours on a single NVIDIA RTX 2080 TI. Training our model on the largest real image dataset (FFHQ) at $256 \times 256$ resolution takes approximately $4$ days using two NVIDIA V100 GPU.

\section{Semi-Supervised Disentanglement without Residual Attributes}
\label{app:semi_no_res}
We evaluate our method on the synthetic datasets with all the factors of variation treated as attributes of interest, holding out no residual factors at all. This is the setting studied by \citet{locatello2020fewlabels}. While this is not the task we aim to solve in this paper, our method performs better than \cite{locatello2020fewlabels}, using the same beta-VAE based architecture, as can be seen in Tab.~\ref{tab:semi_no_res}. This highlights the advantage of latent optimization for disentanglement as discussed in \cite{gabbay2020lord}.

\begin{table}
  \caption{Evaluation on synthetic benchmarks in the setting where there are no residual attributes \cite{locatello2020fewlabels}, using 1000 labels per attribute of interest (mean [std]).}
  \label{tab:semi_no_res}
  \centering
  \begin{tabular}{lllll}
    \toprule
    
        &  &	DCI Disentanglement				&	SAP & MIG				\\
    \midrule
    
\multirow{2}{*}{Shapes3D}	 &	 Locatello \cite{locatello2020fewlabels}	&	0.99	[0.001]	&	0.23	[0.01] 	&	0.75	[0.05]   	\\
& Ours	&	\textbf{1.00}	[0.001]	&	\textbf{0.37}	[0.001] 	&	\textbf{0.99}	[0.01] 	\\

\midrule
											
\multirow{2}{*}{Cars3D}	&	 Locatello \cite{locatello2020fewlabels}	&	0.58	[0.05]	&	0.14	[0.01] 	&	0.25	[0.01]   	\\
& Ours	&	\textbf{0.59}	[0.06]	&	\textbf{0.19}	[0.01] 	&	\textbf{0.49}	[0.01] 	\\
											
\midrule

\multirow{2}{*}{dSprites}	&	 Locatello \cite{locatello2020fewlabels}	&	0.46	[0.03]	&	0.07	[0.001] 	&	0.33	[0.01]   	\\
& Ours	&	\textbf{0.62}	[0.01]	&	\textbf{0.09}	[0.01] 	&	\textbf{0.39}	[0.01] 	\\

\midrule

\multirow{2}{*}{SmallNORB}	&	 Locatello \cite{locatello2020fewlabels}	&	0.43	[0.02]	&	0.13	[0.01] 	&	0.24	[0.01]   	\\
& Ours	&	\textbf{0.68}	[0.01]	&	\textbf{0.31}	[0.22] 	&	\textbf{0.52}	[0.01] 	\\

    \bottomrule
  \end{tabular}
\end{table}

\section{Zero-shot Labeling with CLIP}

\subsection{Implementation Details}
We will provide here a comprehensive description of our method to annotate given images according to the attributes supplied by the user, utilizing the CLIP \cite{radford2021clip} network. For the annotation, the user provides a list of \textit{attributes}, and for each attribute, a list of possible values indexed by $w \in [m^j]$: $s_j^w$ (see Sec.~\ref{sec:att_tables}). For each attribute $j$ and every possible value index $w$, we infer its embedding $u_j^w \in \mathbb{R}^{512}$ using language embedding head of the CLIP model $\phi_{lang}$:
\begin{equation}
\label{eq:language_emb}
u_j^w = \phi_{lang}(s_j^w)
\end{equation}
To obtain a similar embedding for images, we pass each image $x_1,x_2,...,x_n \in \mathcal{X}$ through the vision-transformer (ViT) head of the clip model $\phi_{vis}$. We obtain the representation of each image in the joint embedding space $v_i \in \mathbb{R}^{512}$:
\begin{equation}
\label{eq:language_emb}
v_i = \phi_{vis}(x_i)
\end{equation}
We are now set to assign for each image $i$ and each attribute $j$ their assignment value $a_{ij}$ (one of the $[m^j]$ possible values, or alternatively, the value "$-1$"). For each image we set the value $w \in [m^j]$ to $a_{ij}$, if the image embedding $v_i$ is among the top $K$ similar images to the value embedding $u_j^w$ in the cosine similarity metric (noted by $d$). With a slight abuse of notation:
\begin{equation}
\label{eq:language_emb}
a_{ij} = \{w~ | ~ \big|\{ l~ |  ~d(u_j^w,v_l) \leq d(u_j^w,v_i) \}\big| \leq K \}
\end{equation}
where $\big|\cdot\big|$ is the number of elements in a set.

The value "$-1$" is assigned to $a_{ij}$ in cases where our zero shot classification deem that image uncertain: if an image not among the top $K$ matches for any of the values $u_j^w$, or if it is among the top $K$ matches for more than one of them. 

A high value of $K$ indicates that many images will be labeled for each value $u_j^w$, resulting in a more extensive, but sometime noisy supervision. Many images are not mapped in a close proximity to the embedding of all the sentences describing them, or are not described by any of the user supplied values. Therefore, we would not like to use very large values for $K$. A low value of $K$ sets a more limited supervision, but with more accurate labels. The value of $K$ chosen in practice is indicated in the tables in Sec.~\ref{sec:att_tables}.

\subsection{Attribute Tables}
\label{sec:att_tables}
We include the entire list of the attributes and their possible values for each of the datasets annotated with CLIP. These values are used by our method, and by the LORD \cite{gabbay2020lord} baseline. The lists were obtained as follows. \textbf{FFHQ}: We aggregated human face descriptors, similar to the ones used by other methods, to obtain the attributes in Tab.~\ref{tab:ffhq_att}. \textbf{AFHQ}: To obtain candidate values for animal species we used an online field guide. We randomly selected $200$ images of the AFHQ dataset and identified them to obtain the list in Tab.~\ref{tab:afhq_att}. \textbf{Cars}: We used a few cars types and colors as described in Tab.~\ref{tab:cars_att}.

For comparison with competing methods we used the attribute descriptions brought in their cited paper, or in the code published by the authors. If no similar attribute appeared in the competing methods, we tried a few short descriptions of the attribute, similar to the ones used by our method.

\section{Datasets}
\textbf{FFHQ \cite{karras2019stylegan1}} $70,000$ high-quality images containing considerable variation in terms of age, ethnicity and image background. We use the images at $256 \times 256$ resolution. We follow \cite{karras2020stylegan2} and use all the images for training. The images used for the qualitative visualizations contain random images from the web and samples from CelebA-HQ. 

\textbf{AFHQ \cite{choi2020stargan}} $15,000$ high quality images categorized into three domains: cat, dog and wildlife. We use the images at $128 \times 128$ resolution, holding out 500 images from each domain for testing.

\textbf{Cars \cite{carsdataset}} $16,185$ images of 196 classes of cars. The data is split into $8,144$ training images and $8,041$ testing images. We crop the images according to supplied bounding boxes, and resize the images to $128  \times  128$ resolution (using "border reflect" to avoid distorting the image due to aspect ratio changes).

\section{Extended Qualitative Visualizations}

\clearpage

\begin{table}[H]
\centering
\caption{Values for each attribute used in ZeroDIM for the FFHQ dataset.}
\label{tab:ffhq_att}
\begin{tabular}{ccc}
    \toprule
    Attribute  & Values ($K = 1000$)  \\
    \midrule
age & a kid, a teenager, an adult, an old person \\ 
gender & a male, a female \\ 
ethnicity & a black person, a white person, an asian person \\ 
hair & \{brunette, blond, red, white, black\} hair, bald \\ 
makeup & makeup, without makeup \\ 
beard & a person with a \{beard, mustache, goatee\}, a shaved person \\ 
glasses & a person with glasses, a person with shades, a person without glasses \\ 

	\bottomrule
\end{tabular}
\end{table}

\begin{table}[h!]
\centering
\caption{Values for each attribute used in ZeroDIM for the AFHQ dataset.}
\label{tab:afhq_att}
\begin{tabular}{ccc}
    \toprule
    Attribute  & Values ($K = 100$)  \\
    \midrule
\multirow{12}{*}{species} & \multicolumn{1}{l}{russell terrier, australian shepherd, caucasian shepherd, boerboel,  golden retriever, } \\
& \multicolumn{1}{l}{labradoodle, english foxhound, shiba inu, english shepherd, saluki, husky,} \\
& \multicolumn{1}{l}{flat-coated retriever, charles spaniel, chihuahua, dalmatian, cane corso, bengal tiger,}  \\
& \multicolumn{1}{l}{sumatran tiger, german shepherd, carolina dog, irish terrier, usa shorthair, lion, }  \\ 
& \multicolumn{1}{l}{snow leopard, lion cat, british shorthair,  bull terrier, welsh ke ji, cheetah of asia,} \\
& \multicolumn{1}{l}{himalayan cat, shetland sheepdog, egyptian cat, bombay cat, american bobtail, } \\ & \multicolumn{1}{l}{labrador, american wirehair, chinese li hua, chinese kunming dog, snowshoe cat, } \\ & \multicolumn{1}{l}{maine cat, arctic fox, norwegian forest cat, king shepherd,  beagle, ragdoll,  } \\  
& \multicolumn{1}{l}{brittany hound, tricolor cat, border collie, stafford bull terrier, ground flycatcher,} \\
& \multicolumn{1}{l}{manchester terrier, entrebuche mountain dog, poodle, west highland white terrier,} \\
& \multicolumn{1}{l}{chesapeake bay retriever, hofwald, weimaraner, samoye, hawksbill cat, grey wolf,} \\
& \multicolumn{1}{l}{grey fox, lionesses, singapore cat,  african wild dog, yorkshire dog, persian leopard,} \\
& \multicolumn{1}{l}{stacy howler, hygen hound, european shorthair, farenie dog, siberia tiger, jaguar} \\

	\bottomrule
\end{tabular}
\end{table}

\begin{table}[h!]
\centering
\caption{Values for each attribute used in ZeroDIM for the Cars dataset.}
\label{tab:cars_att}
\begin{tabular}{ccc}
    \toprule
    Attribute  & Values ($K = 500$)  \\
    \midrule
\multirow{1}{*}{type}  & a jeep, a sports car, a family car \\ 
\multirow{1}{*}{color}    & a \{black, white, blue, red, yellow\} car \\ 

	\bottomrule
\end{tabular}
\end{table}

\begin{table}[h!]
\centering
\caption{betaVAE architecture of the generator $G$ in our synthetic experiments. }
\label{tab:generator_betavae}
\begin{tabular}{ccccc}
    \toprule
    Layer & Kernel Size & Stride & Activation & Output Shape \\
    \midrule
    Input & - & - & - & $10$ \\
    FC & - & - & ReLU & 256 \\
    FC & - & - & ReLU & $4 \times 4 \times 64 $ \\
    ConvTranspose & $4 \times 4$ & $2 \times 2$ & ReLU & $8 \times 8 \times 64$ \\
    ConvTranspose & $4 \times 4$ & $2 \times 2$ & ReLU & $16 \times 16 \times 32$ \\
    ConvTranspose & $4 \times 4$ & $2 \times 2$ & ReLU & $32 \times 32 \times 32$ \\
    ConvTranspose & $4 \times 4$ & $2 \times 2$ & Sigmoid & $64 \times 64 \times$ channels \\

	\bottomrule
\end{tabular}
\end{table}

\begin{table}[h!]
\centering
\caption{betaVAE architecture of the classifiers $C^j$ and the residual encoder $E_r$ in our synthetic experiments. $D$ is set to the number of classes $m^j$ of attribute $j$ or the residual dimension $10 - k$.}
\label{tab:encoder_betavae}
\begin{tabular}{ccccc}
    \toprule
    Layer & Kernel Size & Stride & Activation & Output Shape \\
    \midrule
    Input & - & - & - & $64 \times 64 \times$ channels \\
    Conv & $4 \times 4$ & $2 \times 2$ & ReLU & $32 \times 32 \times 32$ \\
    Conv & $4 \times 4$ & $2 \times 2$ & ReLU & $16 \times 16 \times 32$ \\
    Conv & $2 \times 2$ & $2 \times 2$ & ReLU & $8 \times 8 \times 64$ \\
    Conv & $2 \times 2$ & $2 \times 2$ & ReLU & $4 \times 4 \times 64$ \\
    FC & - & - & ReLU & 256 \\
    FC & - & - & - & $D$ \\
	\bottomrule
\end{tabular}
\end{table}

\begin{table}[h!]
\centering
\caption{StyleGAN2-based generator architecture in our experiments on real images. StyleConv and ModulatedConv use the injected latent code which is a concatenation of the representations of the attributes of interest and the residual attributes.}
\label{tab:generator_stylegan2}
\begin{tabular}{ccccc}
    \toprule
    Layer & Kernel Size & Activation & Resample & Output Shape \\
    \midrule
    Constant Input & - & - & - & $4 \times 4 \times 512$ \\
    StyledConv & $3 \times 3$ & FusedLeakyReLU & - & $4 \times 4 \times 512$ \\
    StyledConv & $3 \times 3$ & FusedLeakyReLU & UpFirDn2d & $8 \times 8 \times 512$ \\
    StyledConv & $3 \times 3$ & FusedLeakyReLU & - & $8 \times 8 \times 512$ \\
    StyledConv & $3 \times 3$ & FusedLeakyReLU & UpFirDn2d & $16 \times 16 \times 512$ \\
    StyledConv & $3 \times 3$ & FusedLeakyReLU & - & $16 \times 16 \times 512$ \\
    StyledConv & $3 \times 3$ & FusedLeakyReLU & UpFirDn2d & $32 \times 32 \times 512$ \\
    StyledConv & $3 \times 3$ & FusedLeakyReLU & - & $32 \times 32 \times 512$ \\
    StyledConv & $3 \times 3$ & FusedLeakyReLU & UpFirDn2d & $64 \times 64 \times 512$ \\
    StyledConv & $3 \times 3$ & FusedLeakyReLU & - & $64 \times 64 \times 512$ \\
    StyledConv & $3 \times 3$ & FusedLeakyReLU & UpFirDn2d & $128 \times 128 \times 256$ \\
    StyledConv & $3 \times 3$ & FusedLeakyReLU & - & $128 \times 128 \times 256$ \\
    StyledConv & $3 \times 3$ & FusedLeakyReLU & UpFirDn2d & $256 \times 256 \times 128$ \\
    StyledConv & $3 \times 3$ & FusedLeakyReLU & - & $256 \times 256 \times 128$ \\
    ModulatedConv & $1 \times 1$ & - & - & $256 \times 256 \times 3$ \\
	\bottomrule
\end{tabular}
\end{table}

\begin{table}[H]
\centering
\caption{StyleGAN2-based discriminator architecture in our experiments on real images.}
\label{tab:discriminator_stylegan2}
\begin{tabular}{ccccc}
    \toprule
    Layer & Kernel Size & Activation & Resample & Output Shape \\
    \midrule
    Input & - & - & - & $256 \times 256 \times 3$ \\
    Conv & $3 \times 3$ & FusedLeakyReLU & - & $256 \times 256 \times 128$ \\
    ResBlock & $3 \times 3$ & FusedLeakyReLU & UpFirDn2d & $128 \times 128 \times 256$ \\
    ResBlock & $3 \times 3$ & FusedLeakyReLU & UpFirDn2d & $64 \times 64 \times 512$ \\
    ResBlock & $3 \times 3$ & FusedLeakyReLU & UpFirDn2d & $32 \times 32 \times 512$ \\
    ResBlock & $3 \times 3$ & FusedLeakyReLU & UpFirDn2d & $16 \times 16 \times 512$ \\
    ResBlock & $3 \times 3$ & FusedLeakyReLU & UpFirDn2d & $8 \times 8 \times 512$ \\
    ResBlock & $3 \times 3$ & FusedLeakyReLU & UpFirDn2d & $4 \times 4 \times 512$ \\
    Concat stddev & $3 \times 3$ & FusedLeakyReLU & UpFirDn2d & $4 \times 4 \times 513$ \\
    Conv & $3 \times 3$ & FusedLeakyReLU & - & $4 \times 4 \times 512$ \\
    Reshape & - & - & - & 8192 \\
    FC & - & FusedLeakyReLU & - & 512 \\
    FC & - & - & - & 1 \\
    
	\bottomrule
\end{tabular}
\end{table}

\begin{table}[H]
\centering
\caption{StarGAN-v2-based encoder architecture for the residual attributes in our experiments on real images.}
\label{tab:encoder_stargan2}
\begin{tabular}{ccccc}
    \toprule
    Layer & Kernel Size & Activation & Resample & Output Shape \\
    \midrule
    Input & - & - & - & $256 \times 256 \times 3$ \\
    Conv & $3 \times 3$ & - & - & $256 \times 256 \times 64$ \\
    ResBlock & $3 \times 3$ & LeakyReLU ($\alpha = 0.2$) & Avg Pool & $128 \times 128 \times 128$ \\
    ResBlock & $3 \times 3$ & LeakyReLU ($\alpha = 0.2$) & Avg Pool & $64 \times 64 \times 256$ \\
    ResBlock & $3 \times 3$ & LeakyReLU ($\alpha = 0.2$) & Avg Pool & $32 \times 32 \times 256$ \\
    ResBlock & $3 \times 3$ & LeakyReLU ($\alpha = 0.2$) & Avg Pool & $16 \times 16 \times 256$ \\
    ResBlock & $3 \times 3$ & LeakyReLU ($\alpha = 0.2$) & Avg Pool & $8 \times 8 \times 256$ \\
    ResBlock & $3 \times 3$ & LeakyReLU ($\alpha = 0.2$) & Avg Pool & $4 \times 4 \times 256$ \\
    
    Conv & $4 \times 4$ & LeakyReLU ($\alpha = 0.2$) & - & $1 \times 1 \times 256$ \\
    Reshape & - & - & - & 256 \\
    FC & - & - & - & $D$ \\
	\bottomrule
\end{tabular}
\end{table}

\section{License of Used Assets}
The assets CLIP \cite{radford2021clip}, TediGAN \cite{xia2021towards} and StyleCLIP \cite{patashnik2021styleclip} use the 'MIT License'.

StyleGAN2 \cite{karras2020stylegan2} uses the 'Nvidia Source Code License-NC'.

\end{document}